\documentclass[lettersize,journal,preprint]{IEEEtran}

\usepackage{url}

\usepackage{array}
\usepackage{textcomp}
\usepackage{stfloats}
\usepackage{verbatim}
\usepackage{cite}
\usepackage{caption}
\usepackage{subcaption}
\usepackage{algorithmic}
\usepackage{algorithm}
\usepackage{graphicx}
\usepackage{amsmath,amsfonts}
\usepackage{amssymb}
\usepackage{booktabs}
\usepackage[pagebackref=true,breaklinks=true,colorlinks,bookmarks=false]{hyperref}

\usepackage[capitalize]{cleveref}
\crefname{section}{Sec.}{Secs.}
\Crefname{section}{Section}{Sections}
\Crefname{table}{Table}{Tables}
\crefname{table}{Tab.}{Tabs.}

\usepackage{orcidlink}

\usepackage{epsfig}

\usepackage{microtype}
\usepackage{subcaption}
\usepackage{xcolor}

\usepackage[utf8]{inputenc}
\usepackage{multirow}
\usepackage{outlines}
\usepackage{tikz-cd}

\newcommand{\lmse}{\calL_\text{MSE}}
\newcommand{\lml}{\calL_\text{ML}}

\newcommand{\oneb}{{\mathbf 1}}

\newcommand{\calL}{\mathcal{L}}

\usepackage{amsmath,amsfonts,bm}

\def\1{\bm{1}}

\DeclareMathAlphabet{\mathsfit}{\encodingdefault}{\sfdefault}{m}{sl}
\SetMathAlphabet{\mathsfit}{bold}{\encodingdefault}{\sfdefault}{bx}{n}

\newcommand{\E}{\mathbb{E}}

\newcommand{\R}{\mathbb{R}}

\DeclareMathOperator*{\argmax}{arg\,max}
\DeclareMathOperator*{\argmin}{arg\,min}
\DeclareMathOperator{\EX}{\mathbb{E}}

\newcommand{\map}{$g^{\dagger}$-MAP}
\newcommand{\rev}[1]{{\color{black}#1}}
\newcommand{\secrev}[1]{{\color{black}#1}}

\begin{document}

\title{Conditional Injective Flows for Bayesian Imaging}

\author{AmirEhsan Khorashadizadeh$^{\orcidlink{0000-0003-0660-6823}}$, Konik Kothari, Leonardo Salsi, Ali Aghababaei Harandi, Maarten de Hoop$^{\orcidlink{0000-0002-6333-0379}}$  and Ivan Dokmani\'c$^{\orcidlink{0000-0001-7132-5214}}$
\thanks{
Manuscript received April 18, 2022; revised September 5, 2022 and January
1, 2023; accepted February 9, 2023.\\ 
AmirEhsan Khorashadizadeh and Ivan Dokmani\'c were supported by the European Research Council Starting Grant 852821--SWING. Maarten de Hoop gratefully acknowledges support from the Department of Energy under grant DE-SC0020345, the Simons Foundation under the MATH + X program, and the corporate members of the Geo-Mathematical Imaging Group at Rice University. (Corresponding author: AmirEhsan Khorashadizadeh.)

AmirEhsan Khorashadizadeh and Leonardo Salsi are with the Department of Mathematics and Computer Science of the University of Basel, 4001 Basel, Switzerland (e-mail: \href{mailto:amir.kh@unibas.ch}{amir.kh@unibas.ch}, \href{mailto:leonardo.salsi@outlook.com}{leonardo.salsi@outlook.com}).

Konik Kothari is with the Department of Electrical and Computer Engineering, the University of Illinois at Urbana-Champaign, Urbana, IL 61801 USA (e-mail: \href{mailto:kkothar3@illinois.edu}{kkothar3@illinois.edu}).

Ali Aghababaei Harandi is with the Department of Electrical Engineering, Sharif University of Technology, Tehran 14399-57131, Iran (e-mail: \href{mailto:aghababaei.ali94@gmail.com}{aghababaei.ali94@gmail.com}).

Maarten V. de Hoop is with the Department of Computational and Applied
Mathematics, Rice University, Houston, TX 77005 USA
(e-mail: \href{mailto:mdehoop@rice.edu}{mdehoop@rice.edu}).

Ivan Dokmani\'c is with the Department of Mathematics and Computer Science of the University of Basel, 4001 Basel, Switzerland, and also with the Department of Electrical, Computer Engineering, the University of Illinois at Urbana-Champaign, Urbana, IL 61801 USA (e-mail: \href{mailto: ivan.dokmanic@unibas.ch}{ivan.dokmanic@unibas.ch}). \\
Our implementation is available at \url{https://github.com/swing-research/conditional-trumpets}.
}
}

\maketitle

\begin{abstract}
Most deep learning models for computational imaging regress a single reconstructed image. 
In practice, however, ill-posedness, nonlinearity, model mismatch, and noise often conspire to make such \textit{point estimates} misleading or insufficient. The Bayesian approach models images and (noisy) measurements as jointly distributed random vectors and aims to approximate the posterior distribution of unknowns.
Recent variational inference methods based on conditional normalizing flows are a promising alternative to traditional MCMC methods, but they come with drawbacks: excessive memory and compute demands for moderate to high resolution images and underwhelming performance on hard nonlinear problems.
In this work, we propose C-Trumpets---conditional injective flows specifically designed for imaging problems, which greatly diminish these challenges. Injectivity reduces memory footprint and training time while low-dimensional latent space together with architectural innovations like fixed-volume-change layers and skip-connection revnet layers, C-Trumpets outperform regular conditional flow models on a variety of imaging and image restoration tasks, including limited-view CT and nonlinear inverse scattering, with a lower compute and memory budget. C-Trumpets enable fast approximation of point estimates like MMSE or MAP as well as physically-meaningful uncertainty quantification.
\end{abstract}

\begin{IEEEkeywords}
Bayesian Imaging, Amortized Variational Inference, Deep Generative Models, Computed Tomography, Inverse Scattering, Normalizing Flows
\end{IEEEkeywords}

\section{Introduction}
In Bayesian modeling of computational imaging problems, we assume that the (unknown) object of interest $x$ and the observed measurements $y$ are realizations of random vectors $X \in \mathcal{X}$ and $Y \in \mathcal{Y}$ with  a joint distribution $p_{X, Y}$. This general model includes the common setting of a deterministic forward operator and additive Gaussian noise,
\begin{equation}
y = A(x) + \epsilon,
\label{eq:inverse_prob}
\end{equation}
where
\begin{equation}
Y|X \sim \mathcal{N}(A(X),\sigma^2 I)
\label{eq:likelihood model}
\end{equation}
as well as other relevant models like $Y = \mathrm{Poisson}(A(X), \lambda)$ or even random or uncertain forward operators $A$.

Common machine-learning approaches to  \textbf{ill-posed} inverse problems yield point estimates, that is, they output a single reconstruction. For example, training a deep neural network $f_\theta$ with the mean-squared error (MSE) loss $\E ~ \|X - f_\theta(Y)\|^2$ approximates the minimum-mean-squared-error (MMSE) estimator\footnote{Often called the regression function in machine learning and statistics.} $\E[X | Y]$ (the posterior mean) \cite{jin2017deep}.

In many situations, however, a single point estimate can be misleading or incomplete.
For example, in radio interferometric imaging which aims to reconstruct astronomical images from radio telescope measurements, there can be multiple solutions that fit the observed measurements; a now-famous example is the imaging of a black hole~\cite{sun2020deep}. This can happen for a variety of reasons, all stemming from the ill-posedness of the imaging problem.
The posterior may be multimodal, in which case the MMSE estimator blends the modes together and maximum a posteriori estimate  (MAP) (or posterior mode) $\argmax_x p_{X|Y}(x|y)$, returns only one of the many modes. Even when the posterior is unimodal, providing a point estimate when measurements have low signal-to-noise ratio does not convey the amount of uncertainty in the estimate, thus requiring cautious interpretation. As a remedy, uncertainty quantification (UQ) on top of a reconstructed image can greatly help medical professionals make more informed decisions or order additional measurements in uncertain regions~\cite{vasconcelos2022uncertainr}. 

\begin{figure}
\centering
\includegraphics[width=0.9\linewidth]{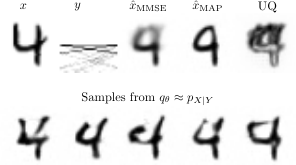}
\caption{Travel-time tomography with boundary sensors; \rev{we measure the wave travel time between sensors in Figure~\ref{fig:setup travel-time}}. The column $x$ shows the ground truth, the column $y$ the pseudo-inverse of the forward operator applied to the measurements. Conditional injective flows provide MMSE, surrogate MAP (Section \ref{sec:fixed_vol_change_layers}) and physically meaningful uncertainty quantification (UQ). Given no sensors in the top regions of the image, the trained C-Trumpet assigns higher uncertainty to the top half of the domain. While both standard point estimates look close to ``9'', many posterior samples look like ``4''s which indicates the significance of posterior sampling for ill-posed inverse problems.}
\label{fig:intro_tt}
\end{figure}

In Figure~\ref{fig:intro_tt} we use a toy problem to illustrate access to the posterior can help in the presence of multiple modes. Both standard point estimates look close to a ``9'', but many posterior samples look like ``4''s. While innocuous on MNIST~\cite{lecun1998gradient}, such failure modes come with risks in medical imaging.

Another approach is then to characterize the posterior
$p_{X| Y}$ upon observing $y$.
Given the joint distribution $p_{X, Y}$, computing the posterior $p_{X|Y}$ generally involves intractable high-dimensional integrations. A standard way to circumvent this is by sampling methods such as Markov chain Monte Carlo (MCMC) \cite{chen2012monte,bishop2006pattern}. These methods work well in low-dimensional problems but become computationally intensive when used in high-dimensional imaging tasks due to the need to compute the forward operator many times~\cite{pmlr-v80-marino18a}. An alternative is to perform \textit{variational inference}. This requires defining a tractable, parameterized family of distributions $\mathcal{Q}$ and choosing a $q \in \mathcal{Q}$ that is ``close'' to $p_{X|Y}$. 

In this paper, we define a new class of approximate posteriors $\mathcal{Q}$ using injective deep neural networks, suitable for imaging problems. 
We build on the recently-proposed injective flows \cite{kumar2020regularized,brehmer2020flows,kothari2021trumpets} and conditional coupling layers~\cite{ardizzone2021conditional}.
Injective flows map a low-dimensional latent space to a high-dimensional image space using a sequence of injective functions whose inverses (on the range) can be computed efficiently and exactly. They combine the favorable aspects of GANs \cite{mirza2014conditional} and invertible normalizing flows~\cite{dinh2014nice,dinh2016density,kingma2018glow} in a way especially well-suited for imaging problems. \rev{Indeed, modern GANs with the various innovations in architectures and training protocols generate subjectively high-quality high-resolution images~\cite{karras2019style,karras2020analyzing} and can be trained relatively straightforwardly with a reasonable compute budget} but lack an exact, fast inverse and access to (approximate) likelihoods. Normalizing flows have fast inverses and enable fast and exact likelihood computation \rev{(although this depends on the particular architecture)}, but they lack a low-dimensional latent space thereby requiring large memory and compute budget when training at higher image resolutions~\cite{yu2020wavelet}. Injective flows have a low-dimensional latent space with a fast and exact inverse on the range while maintaining a low compute and memory budget.

\emph{Conditional} normalizing flows~\cite{ardizzone2021conditional} inherit the favorable properties of their non-conditional counterparts---easy access to the likelihoods and inverses of generated samples. They were applied to inverse problems~\cite{winkler2019learning,padmanabha2021solving} where they enable posterior estimation, efficient sampling and even uncertainty quantification. In this work, we propose \textbf{conditional injective flows} termed C-Trumpets. While injective flows model image datasets supported on low-dimensional manifolds, the range of a C-Trumpet is a (potentially) different low-dimensional manifold of posterior samples \emph{for each}  measurement. As we show in Section \ref{sec: fiber bundle}, this makes them an effective model for data distributions supported on \emph{fiber bundles} with the base space corresponding to the space of measurements $\mathcal{Y}$, and in particular  effective models of posterior distributions in imaging inverse problems. Moreover, thanks to a low-dimensional latent space, they can be trained for high-dimensional data ($256 \times 256$) using a single GPU while training conditional bijective flows at this resolution requires significantly more memory.

While generating approximate posterior samples is important, many applications still call for Bayesian point estimates such as the MAP or the MMSE estimator, and it is convenient if those can be computed with the same model. While it is clear (at least conceptually) how to do it for the MMSE estimator---generate many samples and average them---the MAP estimator is more elusive. There are not many deep learning approaches to imaging which compute (or approximate) the MAP estimator since the corresponding training loss (purely 
formally) is the highly irregular $\delta(x - x')$ as opposed to the ``nice'' $\| x - x' \|^2$~\cite{gribonval2011should}. (A notable exception is amortized MAP for image super-resolution \cite{sonderby2016amortised}, although it is limited to noiseless linear low-rank projections.) We could in principle obtain a MAP estimate from C-Trumpets via iterative maximization; however, that is slow and it is not guaranteed to converge. We thus propose a modification of coupling layers which results in a fixed volume change with respect to the input. We use this newly designed layer to efficiently approximate MAP estimators without the need to run an iterative process or evaluate the forward operator.

Our main contributions can be summarized as follows:
\begin{itemize}
    \item We define a class of deep learning models for amortized variational inference called C-Trumpets, with a smaller memory and compute footprint compared to bijective flows; C-Trumpets can be trained on $256 \times 256$ images on a single V100 GPU in a day.
    \item The new flows include bespoke architectural innovations: fixed-volume-change layers provide efficient MAP estimates without iterative optimization, while skip
    connections improve the quality of the generated samples and uncertainty quantification.
    \item We show that  C-Trumpets outperform conditional bijective flows in solving computational imaging problems including nonlinear electromagnetic scattering, limited-view CT and seismic travel-time tomography; C-Trumpets produce much better posterior samples and uncertainty estimates that are consistent with the physics of the forward operators in various inverse problems.
    
    \item While standard injective flows parameterize manifolds and are thus a natural (regularizing) choice when we believe the manifold assumption holds, we show that conditional injective flows can parameterize fiber bundles~\cite{courts2021bundle}.
    
\end{itemize}

\section{Variational Bayesian Inference} \label{sec:variational_posterior}

In this section we work formally and assume that all probability measures have a density; this allows us to simply present the main ideas. A training strategy suitable for distributions supported on low-dimensional manifolds is detailed in Section~\ref{sec:C-Trumpet}.

We consider random vectors $X \in \mathcal{X}$ (representing the unknown object) and $Y \in \mathcal{Y}$ (representing the observed noisy measurements). The posterior distribution $p_{X|Y}$ can be expressed as
\begin{equation}
    p_{X|Y}(x|y) = \dfrac{p_{Y|X}(y|x) p_X(x)}{\int_x p_{X,Y}(x,y)  dx}.
\end{equation}
In high-dimensional imaging applications, calculating $\int_x p_{X,Y}(x,y) dx$ is intractable. Moreover, the prior distribution $p_X$ is unknown and needs to be estimated~\cite{jalal2021robust,bohra2022bayesian}. MCMC-based sampling methods do not require access to $\int_x p_{X,Y}(x,y) dx$ but they are slow if one desires likely posterior samples~\cite{wang2018randomized, pmlr-v80-marino18a}. An alternative to MCMC is to perform variational inference~\cite{hinton1993keeping,graves2011practical}:
we define a parameterized class of distributions 
$$
    \mathcal{Q} = \{ q_\theta \in \mathcal{P}(\mathcal{X}) \ : \ \theta \in \Theta \},
$$
where $\mathcal{P}(\mathcal{X})$ is the space of probability distributions over $\mathcal{X}$, and we search for $\theta$ such that $q_\theta$ is close to $p_{X|Y}$. Examples of $\mathcal{Q}$ include Gaussian mixtures and densities induced by generative neural networks.

The remaining ingredients are a measure of ``closeness'' and a fitting algorithm. 
In standard variational inference it is common to use the Kullback--Leibler (KL) divergence as a measure of fit,
\begin{align*}
    \text{KL}(q \| p) 
    &= \int_\mathcal{X} q(x) \log\left(\dfrac{q(x)}{p(x)}\right) dx\\
    &= \E_{X \sim q}[ \log q(X) -  \log p(X)].
\end{align*}
Given a measurement $y$, we choose $q_\theta$ by solving either
\[
    \theta_{\text{rev}}^\ast(y) = \argmin_{\theta \in \Theta} ~ \text{KL}(q_\theta \| p_{X|Y}( \cdot \, | \, y) ),
\] 
or 
\[
    \theta_{\text{fwd}}^\ast(y) = \argmin_{\theta \in \Theta} ~ \text{KL}(p_{X|Y}( \cdot \, | \, y) \| q_\theta ),
\] 
respectively called reverse and forward KL minimization~\cite{papamakarios2021normalizing,bishop2006pattern}. The two estimates are in general different due to the asymmetry of the KL divergence.

Minimizing the reverse KL requires computing the expectation $\E_{X \sim q_\theta} \log p_{Y|X}(y|X) p_X(X)$. While $p_{Y|X}$ is usually asssumed known in imaging problems with a known forward operator---in~\eqref{eq:likelihood model} it corresponds to additive Gaussian noise---the prior distribution $p_X$ is unknown. Recently, normalizing flows were used to estimate $p_X$ to then allow for downstream minimization of the reverse KL divergence \cite{sun2020deep,whang2021composing,kothari2021trumpets}. On the other hand, minimizing the forward KL does not require access to the prior distribution, the noise model or the forward operator\cite{minka2013expectation,minka2001family}.

Computing $\theta^*(y)$ in both forward and reverse KL formulation involves solving a separate optimization problem for every new measurement $y$. The reason is that $q_\theta \in \mathcal{Q}$ is a function of $x$ alone, not $x$ and $y$, and we hope that for each $y$,
$$
    q_{\theta^*(y)}(x) \approx p_{X|Y}(x|y).
$$
This separate optimization for every $y$ may be inefficient if implemented via standard iterative solvers.

We could, however, work directly with a family of conditional distributions $q_\theta(x|y)$ which depend on both $x$ and $y$,
$$
    \mathcal{Q}_{\text{cond}} = \{ (x, y) \mapsto q_\theta(x|y) \ : \ \theta \in \Theta_{\text{cond}}\},
$$
where for each $y$, $q_\theta( \cdot | y) \in \mathcal{P}(\mathcal{X})$ is a probability distribution over $\mathcal{X}$.

We can now compute a conditional variational approximator $q_\theta(x|y)$
by minimizing the \textit{average} KL divergence over all measurements $y$: this procedure is known as \textit{amortized inference}. It leads to the following optimization problem:
\begin{equation} \label{eq:theta}
\begin{aligned}
    \theta^*
    &= \argmin_{\theta} ~  \mathbb{E}_{Y \sim p_Y}  \text{KL}(p_{X|Y}( \, \cdot\, |Y) \| q_\theta(\, \cdot\, |Y)) \\
    &= \argmax_{\theta} ~ \mathbb{E}_{X, Y \sim p_{X, Y}} \log q_{\theta}(X|Y).
\end{aligned}
\end{equation}
The key observation is that the population expectation over $p_{X, Y}$ can now be approximated by an empirical expectation over the training data~$\{ (x^{(i)}, y^{(i)}) \}_{i = 1}^N$.

The question that remains is: how to parameterize $q_\theta(x|y)$ so that we can 1) learn $\theta^*$ efficiently from data, 2) easily obtain conditional samples from $q_{\theta^*}(x|y) \approx p_{X|Y}(x|y)$ for a given $y$, and 3) efficiently compute standard point estimators such as the MAP estimator? We answer this question in the remainder of the paper by describing conditional injective flows called C-Trumpets.

In a nutshell, we will define a family of neural networks $f_\theta(Z; y)$ where $y$ is the conditioning input. The first argument, $Z$, will be a standard Gaussian random vector over a low-dimensional latent space, and the parameter $\theta$ adjusted so that for each $y$, the random vector $f_\theta(Z; y)$ is close in distribution to $X | Y\!\!=\!\!y$. In other words, denoting the standard Gaussian distribution by $p_Z$, we will obtain $q_\theta(\cdot ~ | ~ y)$ as a pushforward of $p_Z$ via $f_\theta(z; y)$,\footnote{For simplicity we lightly abuse notation by identifying a probability measure and its density.}
\begin{equation} \label{eq:pushforward}
    q_\theta (\cdot ~ | ~ y) = \left[ f_\theta(\,\cdot\,; y) \right]_{\#} p_Z.
\end{equation}
The approximate posterior samples can then be obtained in a standard way as $f_{\theta}(Z;y)$.

\section{C-Trumpets: Conditional Injective Flows}
\label{sec:C-Trumpet}

\begin{figure*}[t!]
\centering
\includegraphics[width=0.9\textwidth]{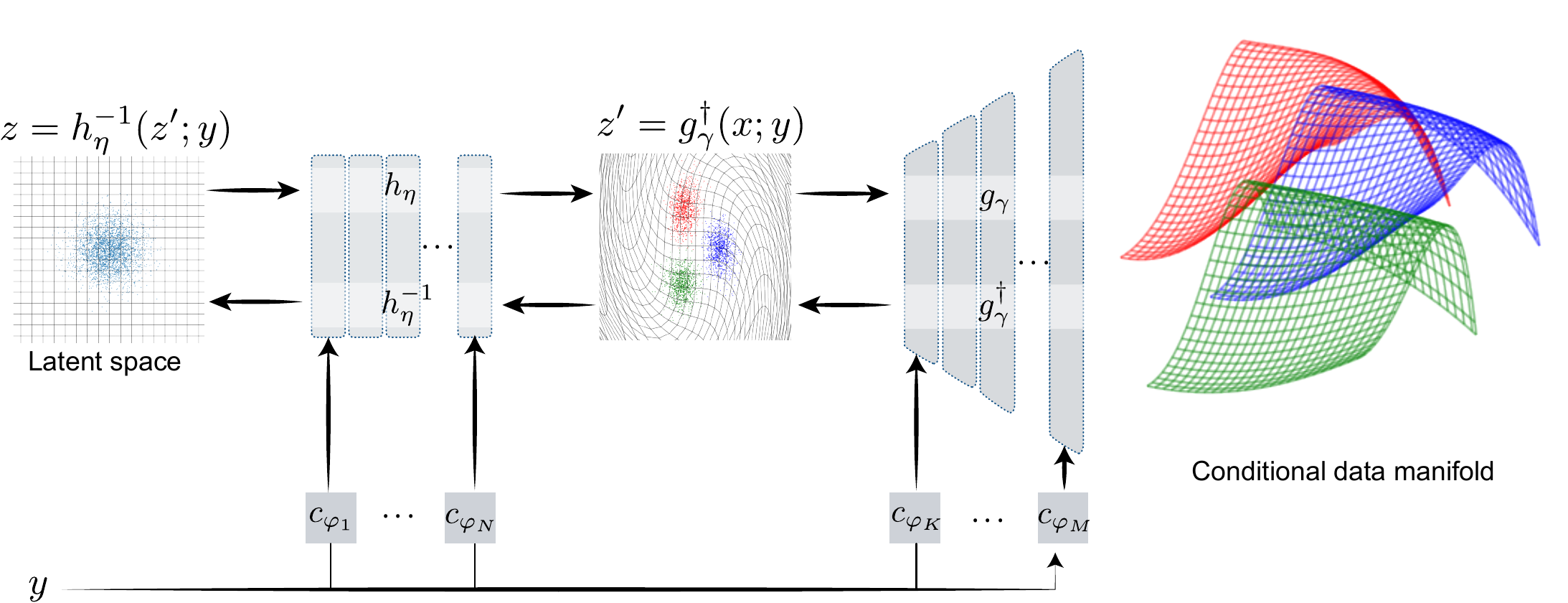}
\caption{Conditional injective flows. Different measurements {\color{red} $y_1$}, {\color{blue} $y_2$} and {\color{green} $y_3$} give different manifolds.}
\label{fig:C-Trumpet}
\end{figure*}

\rev{For many natural and medical image classes, the posterior distribution $p_{X|Y}$ is low-dimensional as its support is a subset of the support of the prior distribution $p_X$ which is assumed to concentrate close to a low-dimensional manifold~\cite{bengio2013representation, brown2022union}}.
C-Trumpets (Figure~\ref{fig:C-Trumpet}) are conditional injective normalizing flows that map a low-dimensional latent space to a high dimensional data space using an injective transformation for each conditioning sample. Injectivity guarantees that for each conditioning sample the range of the network is a manifold. Moreover, the efficiently-invertible layers facilitate training and inference. As shown in Figure~\ref{fig:C-Trumpet}, the proposed model has two subnetworks: an injective generator $g_{\gamma}$ that maps a low-dimensional space $\mathcal{Z} = \R^d$ to the data space $\mathcal{X} = \R^D$, $d \ll D$, and a bijective mapping $h_{\eta}: \mathcal{Z} \to \mathcal{Z}$ maintaining dimensionality. The end-to-end mapping is then given as,
\[
f_\theta(z;y)=g_{\gamma}(h_{\eta}(z;y);y)
\]
with learnable parameters $\theta = (\gamma, \eta)$.

C-Trumpets are inspired by non-conditional injective flows~\cite{kothari2021trumpets} called Trumpets. They comprise a sequence of injective~\cite{kothari2021trumpets} and bijective~\cite{ kingma2018glow} revnet blocks. Each revnet block consists of three components: activation normalization, (injective or bijective) $1 \times 1$ convolution and affine coupling layer. It is worth noting that all these components are non-conditional, and we will later use conditional affine coupling layers to make the generation process conditional.

\begin{enumerate}
    \item activation normalization,
    \begin{equation}
        \begin{array}{ll}
        \text{\textsc{Forward:}} &  ~ x = \dfrac{z-\mu}{\sigma}  \\
            \text{\textsc{Inverse:}} & ~ z = \sigma x + \mu
        \end{array}
    \end{equation}
    \item $1\times1$ convolution with a kernel $w$,
    \begin{enumerate}
        \item Bijective version:
        \begin{equation}
            \begin{array}{ll}
                    \text{\textsc{Forward:}} & ~ x = w\ast z \\  
                    \text{\textsc{Inverse:}} & ~ z = w^{-1}\ast x
            \end{array}
         \end{equation}
        \item Injective version:

        \begin{equation}
            \begin{array}{ll}
                    \text{\textsc{Forward:}} & ~ x = w\ast z \\  
                    \text{\textsc{Inverse:}} & ~ z = w^{\dagger}\ast x
            \end{array}
         \end{equation}
    \end{enumerate}
     $w$ is a $1 \times 1$ convolutional filter, which is simply a matrix multiplication along the channel dimension where $w \in \R^{c_{in} \times c_{out}}$ and $w^\dagger$ is the pseudo-inverse of $w$ (a non-square matrix in the injective dimension-expanding case). \rev{We note that $w$ should obey the appropriate constraints to guarantee injectivity~\cite{puthawala2020globally}.}
     We use LU decomposition in matrix inversion, which significantly reduces training time of the injective part of C-Trumpets (see Section \ref{sec: faster MSE} for more details).
    \item affine coupling layer
    \begin{equation*}
    \begin{array}{lll}
        \text{\textsc{Forward:}} & x_1 = z_1, \quad x_2 = s(z_1) \circ z_2 + b(z_2)\\
        \text{\textsc{Inverse:}} &z_1 = x_1  , \quad z_2 = s(x_1)^{-1} \circ (x_2 - b(x_1)),
    \end{array}
    \end{equation*}
    where $z = [z_1, z_2]^T$ and $x = [x_1, x_2]^2$. The mappings $s$ and $b$ are respectively the scale and the shift networks.
\end{enumerate}

In order to make the generative process conditional, we adapt the conditional affine coupling layers proposed in~\cite{ardizzone2021conditional}. Conditional affine coupling layers keep the advantages of a regular flow model---fast inverses and tractable Jacobian computations, while benefiting from the conditioning framework. Since scale $s(\,\cdot\,)$ and shift $b(\,\cdot\,)$ networks are never inverted, we can concatenate the features of the conditioning sample $y$ to their input without losing invertibility and tractable $\log \det$ Jacobian computation. Accordingly, $s(\,\cdot\,)$ and $b(\,\cdot\,)$ are replaced by $s(\,\cdot, c_{\varphi}(y))$ and $b(\,\cdot,c_{\varphi}(y))$,
\begin{equation*}
\begin{array}{lll}
    \text{\textsc{Forward:}} & x_1 = z_1  \\
    & x_2 = s(z_1,c_{\varphi}(y)) \circ z_2 + b(z_2,c_{\varphi}(y)) \\
    \text{\textsc{Inverse:}} &z_1 = x_1 \\
    & z_2 = s(x_1,c_{\varphi}(y))^{-1} \circ (x_2 - b(x_1,c_{\varphi}(y))),
\end{array}
\end{equation*}
    where $z = [z_1, z_2]^T$, $x = [x_1, x_2]^T$ and $c_{\varphi}(\,\cdot\,)$ represents the conditioning network that extracts appropriate features from $y$. We deploy conditional affine coupling layers in both injective and bijective subnetworks of C-Trumpets. 

\subsection{Conditioning network}
\label{sec: conditioning network}
The role of the conditioning network $c_\varphi(\,\cdot\,)$ is to extract features from conditioning samples $y$ to be used by the affine coupling layers. The architecture of the conditioning network depends on the nature of the conditioning samples. When $y$ is structured as an image, we use convolutional layers; when it is an unstructured 1D vector, we use fully connected layers, as for example in class-based image generation (see Section~\ref{app: class-based image generaation} in the supplemental materials) where the conditioning data are one-hot class encodings. The output dimension of the conditioning network is set to match the input dimension of the scale and shift modules of the coupling layers. The weights of the conditioning networks are trained jointly with the remaining parameters in C-Trumpets via back-propagation using paired training data. Further details about conditioning networks in all numerical experiments are given in Sections~\ref{app: limited-view CT} and~\ref{app: class-based image generaation} in the supplemental materials.

\subsection{Skip connections}
\label{sec:skip connections}

The fact that we use expanding revnet layers allows us to augment the architecture with elements that are not compatible with the standard injective layers.
The conditioning samples often have a pixel-wise dependency on target signals. For example, in image inpaiting with a fixed mask location, the out-of-mask pixels should be simply forwarded to the output. In order to not re-learn these features, we introduce skip connections after every revnet block between the measurements, $y$ and the different resolution levels of the injective subnetwork of C-Trumpets,
\begin{equation}
    \begin{array}{ll}
    \text{\textsc{Forward:}} & ~ x = (\oneb - S) \circ \text{rev}(z;y) + S \circ \text{resize}(y) \\[2mm]
    \text{\textsc{Inverse:}}& ~ z = \text{rev}^{-1} \left(\dfrac{x- S \circ \text{resize}(y)}{\oneb - S} ;y \right),
    \end{array}
    \label{eq: skip connection}
\end{equation}
where $\text{rev}(\,\cdot\,)$ is the conditional revnet block, $S$ is a learnable matrix with entries between $0$ and $1$; $S$ adjusts the amount of the direct mixing of the measurements in the generated posterior sample. We empirically find that skip connections help the injective part of C-Trumpets converge faster and they yield better reconstruction in several imaging problems.

\subsection{Fast MAP estimation via fixed volume-change layers}
\label{sec:fixed_vol_change_layers}

MAP estimation traditionally requires an iterative solution of a maximization problem. Deep neural networks for image reconstruction are traditionally trained with an $\ell^p$ loss ($p = 2$ giving the ubiquitous MSE loss and ultimately an approximation of the MMSE estimator $\mathbb{E}[X | Y]$). There are, however, few attempts at using deep neural networks to approximate the MAP estimate in imaging, possibly because the associated ``loss'' would be a tempered distribution $\delta(x - x')$~\cite{gribonval2011should}. While the MAP and the MMSE estimates coincide when $X$ and $Y$ are jointly Gaussian, they are in general different. For posteriors supported on a low-dimensional manifold the MMSE estimate will generally not lie on the manifold. (We show a qualitative manifestation of this effect in Section~\ref{sec: MAP results})

A notable deep-learning approach to amortized MAP inference for image super-resolution has been proposed by S{\o}nderby et al.~\cite{sonderby2016amortised}. However, their method is only applicable for super-resolution in the ideal noise-free scenario. In this section, we propose a new variant of affine coupling layers, which enables us to obtain the MAP estimate instantaneously with a single forward pass of the trained network. This method is exact when used with bijective flows and approximate for injective flows where it computes the MAP estimate of the pre-image samples $z^\prime$ in Figure~\ref{fig:C-Trumpet}.

Consider a trained conditional normalizing flow model $x = f(z;y)$. The MAP estimate can be obtained by solving
\begin{equation}
    \begin{array}{rcl}
    x_{\text{MAP}} & = & \displaystyle \argmax_x ~ \log(p_{X|Y}(x|y)) \\
    & = & \displaystyle \argmax_x ~ \log(p_Z(z)) - \sum_{k = 1}^{K}\log |\det J_{f^k}|,
    \end{array}
    \label{eq: MAP analysis}
\end{equation}
where $z = f^{-1}(x;y)$. In principle, we can run an iterative maximization process to compute $x_{\text{MAP}}$. In general, this may be slow and it is not guaranteed to converge, even with multiple random restarts. 
 
Let us analyze \eqref{eq: MAP analysis} more closely. While the first term, $\log(p_Z(z))$, has the highest value at $z=\mu_z$ (the mean of the Gaussian), the second term has three components: activation normalization, $1 \times 1$ convolution, and the conditional affine coupling layer. The first two components are linear layers with data-independent Jacobians; their $\log\det$s are thus constant with respect to $x$ and can be omitted from \eqref{eq: MAP analysis}. The $\log \det$ of the Jacobian of conditional affine coupling layers is
\begin{equation}
    \log(|\det J(z)|) = \sum_{i=1}^{l} \log(s^i(z_1,c_{\varphi}(y))),
    \label{eq: log-det affine coupling layer}
\end{equation}
where $s^i(\,\cdot\,)$ is the $i$th element of the output of the scale network.
This term is in general data-dependent. In order to make it data-independent, we propose to use the following activation for the scale network,
\begin{equation}
    s_{\text{FVC}}(z,c_{\varphi}(y)) = \exp(\text{softmax}(m(z,c_{\varphi}(y)))),
    \label{eq: data-independent scale network}
\end{equation}
where $m(\,\cdot\,)$ is an arbitrary neural network and $\text{softmax}(x)$ is defined as
\[
\text{softmax}(x)_i = \dfrac{e^{x_i}}{\sum_{j=1}^l e^{x_j}} \quad \text{for} \quad i=1,2,...,l.
\]
Then we have,
\begin{equation}
     \log(|\det J_{\text{FVC}}(z)|) = \sum_{i=1}^{l} \log(s_{\text{FVC}}^i(z_1,c_{\varphi}(y))) = 1.
    \label{eq: data-independent log-det}
\end{equation}
The newly proposed layer has a data-independent $\log\det$ Jacobian, without losing expressivity, as verified empirically in Sections~\ref{sec:computational imaging} and ~\ref{sec:compressed sensing}.

Now all terms in \eqref{eq: MAP analysis} are independent of $x$, so that
\begin{equation}
    \begin{array}{rcl}
    x_{\text{MAP}} & = & \displaystyle\argmax_x ~ \log(p_{X|Y}(x|y)) \\
    & = & \displaystyle\argmax_x ~ \log(p_Z(z)) \\
    & = & f(\mu_z;y);y),
    \end{array}
    \label{eq: data-independent MAP analysis}
\end{equation}
which is to say that the MAP estimate is  obtained simply by feeding the mean of the Gaussian base distribution into the (bijective) flow. 

While the proposed technique is exact for bijective conditional flows, however the $\log \det$ term for conditional \emph{injective} flows  is given as
\begin{multline}
\label{eq:posterior_distribution}
\log p_{X|Y}(x|y) = \log p_Z(f_\theta^{\dagger}(x;y))
\\
- \dfrac{1}{2}\log |\det [J_{f_\theta}(f_\theta^{\dagger}(x;y))^T
 J_{f_\theta}(f_\theta^{\dagger}(x;y))]|.
\end{multline}
 and equation \eqref{eq:posterior_distribution} cannot be decomposed into a sum of the $\log \det$s of its constituent components.
We can nevertheless use this technique to obtain a MAP estimate in the intermediate $z'$-space as in Figure~\ref{fig:C-Trumpet}. Therefore, we obtain a surrogate of the end-to-end MAP estimate and call it surrogate MAP and denote it as {\map}.

\subsection{Training strategy}
\label{sec:training}

Thanks to simple, tractable inverses and $\log\det$s of layer Jacobians, the parameters of normalizing flows can be fitted via maximum likelihood (ML). However, as C-Trumpets have a low-dimensional latent space, the likelihood is only defined in the range of the injective generator. \rev{Since the weights of the networks are initialized randomly, prior to training this range does not contain the targets in the training set and their end-to-end likelihoods are not defined.} We thus split the training of C-Trumpets, $f_\theta(z; y) = g_{\gamma}(h_\eta(z;y);y)$, into two phases, following the method of Brehmer and Cranmer~\cite{brehmer2020flows} for non-conditional injective flows: (1) The MSE training phase where we only fit the trainable parameters $\gamma$ of the injective network with the goal of adjusting the range of $f_\theta$ to contain the training data, and (2) The ML training phase where we optimize the trainable parameters $\eta$ of the low-dimensional bijective part, maximizing the likelihood of the pre-image (through $g_\gamma$) of the training data.\footnote{For simplicity we denote all trainable parameters of the injective network, \emph{including the weights of the conditioning networks and the skip connections} by $\gamma$, and all trainable parameters of the bijective network, \emph{including the weights of the conditioning networks} by $\eta$.} 

In the first phase, we optimize $\gamma$ by minimizing 
\begin{equation} \label{eq:lmse_conditional}
    \lmse(\gamma) = \\
    \dfrac{1}{N} \sum_{i=1}^N \| x^{(i)}- g_{\gamma}(g_{\gamma}^\dagger(x^{(i)};y^{(i)}); y^{(i)})\|_2^2,
\end{equation}
where $\{(x^{(i)}, y^{(i)})\}_{i = 1}^{N}$ are the training samples and $g^\dagger$ is the layer-wise inverse of the injective subnetwork; therefore, the projection operator on the range of $g_{\gamma}(\,\cdot\,; y)$ is given as $P_{g_{\gamma}}(x;y) := g_{\gamma}(g_{\gamma}^\dagger(x;y);y)$.
After training the injective network for a fixed number of epochs, the range of the network approximately contains the training data. 

We now switch to the second phase: maximizing the likelihood of the projected training samples in the intermediate space ($z'$ in Figure \ref{fig:C-Trumpet}), $\{ g_\gamma^\dag(x^{(i)})\}_{i = 1}^N$, by minimizing the following KL divergence over $\eta$ (cf. Section~\ref{sec:variational_posterior}):
\begin{equation} \label{eq:LML_conditional}
    \lml(\eta) =
    \dfrac{1}{N}\sum_{i=1}^N \left(-\log p_Z(z^{(i)}) + \log |\det J_{h_{\eta}}|\right),
\end{equation}
where $z^{(i)} = h_{\eta}^{-1}(g_{\gamma}^\dagger(x^{(i)};y^{(i)});y^{(i)})$ and $p_Z(z)$ is a standard Gaussian distribution in $\mathbb{R}^d$. In summary, this training strategy first ensures that the range of the injective generator ``interpolates''  the training data and then maximizes the intermediate space likelihoods as proxies to the image-space likelihoods. After training, sampling an approximate posterior sample for a given $y$ is performed by sampling a $z$ from a normal distribution and using the forward pass: $x_{\text{gen}} = g_\gamma(h_\eta(z;y);y)$.

\section{The C-Trumpets Signal Model}
\label{sec: fiber bundle}

In this section we briefly discuss the geometric and topological aspects of C-Trumpets. Readers who care mostly about the practical aspects and numerical results may safely skip ahead to Section~\ref{sec:experimental-results}.

The primary motivation for introducing (non-conditional) injective flows is to model data supported on low-dimensional manifolds~\cite{puthawala2020globally,brehmer2020flows,kothari2021trumpets,puthawala2021universal}. It was empirically shown that for common structured datasets these models are indeed simpler, faster to train and generate higher quality samples than globally invertible normalizing flows which maintain the same dimension across all layers. 

On the other hand, we introduce C-Trumpets from the point of view of uncertainty quantification and posterior sampling rather than geometrical and topological considerations about the class of signals it models. It is nevertheless important to make this  class explicit, since this understanding will guide design choices and circumscribe the range of problems in which C-Trumpets are the right tool of choice.

For every (fixed) conditioning sample $y$, a C-Trumpet becomes a conditional flow modeling the corresponding conditional distribution. Thus for every conditioning sample $y$ the range is a low-dimensional manifold 
$$
    \mathcal{M}_y = \{ f_\theta(z; y) \ : \ z \in \mathcal{Z} \}
$$
with topology induced by the topology of the latent space (the support of $p_Z$). The range of a C-Trumpet as a function of $z$ and $y$ then corresponds to the union of the family of these manifolds indexed by $y$,
$$
    \mathcal{R}_\theta
    = 
    \{ f_{\theta}(z; y) \ : \ z \in \mathcal{Z} , y \in \mathcal{Y} \}
    = 
    \bigcup_{y \in \mathcal{Y}} \mathcal{M}_y.
$$
For a given $(y, z)$ let
$$
\beta (y) =
\{ x \ : \ x = f_\theta(z; y), \ z \in \mathcal{Z} \},
$$
denote the set of all possible signals $x$ that could have caused the observation $y$, or in other words the support of the distribution $p_{X|Y=y}$.

Let us now assume a slightly stronger condition than injectivity of $f_\theta$ in the first argument: that the Jacobian (with respect to both $z$ \textit{and} $y$) of $f_\theta$ has full rank $\dim(\mathcal{Z}) + \dim(\mathcal{Y})$. Note that this does not guarantee global injectivity of the map $(z, y) \mapsto f_\theta(z; y)$, but it does imply that there is a (finite) collection of open sets $\{ U_i \}$ (thought of as neighborhoods in $\mathcal{Y}$) which cover $\mathcal{Y}$, $\bigcup_i U_i = \mathcal{Y}$, such that $f_\theta$ is injective on $\mathcal{Z} \times U_i$ for all $i$ and in fact a homeomorphism between $\mathcal{Z} \times U_i$ and $\beta(U_i)$. The signal model we just described is called a \textit{fiber bundle}~\cite{courts2021bundle,choquet1982analysis}.

Formally, a generic fiber bundle comprises sets $(\mathcal{E}, \mathcal{B}, \mathcal{Z})$ called the total space ($\mathcal{E}$), the base space $(\mathcal{B})$, and the (typical) fiber $(\mathcal{Z})$, together with a surjective map $\pi : \mathcal{E} \rightarrow \mathcal{B}$, with the property that for each $b \in \mathcal{B}$, there exists an open neighborhood $b \in U_i \subseteq \mathcal{B}$ on which $\phi: \mathcal{Z} \times U_i \rightarrow \pi^{-1}(U_i)$ is a homeomorphism. The definition can be illustrated by a commutative diagram~\cite{courts2021bundle},
\begin{equation}
\begin{tikzcd}
 \pi^{-1}(U_i) \arrow[d , "\pi"'] \arrow[r, shift right, "\phi^{-1}"'] & \mathcal{Z} \times U_i \arrow[l , shift right, "\phi"'] \arrow[ld , "\text{proj}"] \\
U_i &  
\end{tikzcd}
\end{equation}
where $\text{proj}: \mathcal{Z} \times U_i \to  U_i$. We have the following correspondence between the standard fiber bundle notation and the notation specific to C-Trumpets used in this paper:
\begin{equation*}
\begin{array}{rcl}
    \textbf{C-Trumpets} && \textbf{Fiber bundles} \\[2mm]
    \mathcal{R}_\theta \phantom{ttt} & \longleftrightarrow & \phantom{ttttt} \mathcal{E} \\
    \mathcal{Z} \phantom{ttt} & \longleftrightarrow & \phantom{ttttt} \mathcal{Z} \\
    \mathcal{Y} \phantom{ttt} & \longleftrightarrow & \phantom{ttttt} \mathcal{B} \\
    \beta \phantom{ttt} & \longleftrightarrow & \phantom{ttttt} \pi^{-1} \\
    f_\theta \phantom{ttt} & \longleftrightarrow & \phantom{ttttt} \phi.
\end{array}
\end{equation*}

\rev{Fiber bundles model a variety of situations in imaging, scientific computing, and physics. In CryoEM we can see the 3D rotation group as a bundle of circles over the 2-sphere, which leads to denoising algorithms~\cite{fan2021cryo,debarnot2022manifold}. Symmetries for the forward operator (for example in phase retrieval) naturally split the data space into equivalence classes that can be modeled as fiber bundles. More generally, any many-to-one machine learning task is well-modeled by a fiber bundle~\cite{courts2021bundle}. We also mention natural connections with gauge equivariance and geometric deep learning~\cite{bronstein2021geometric}.

That said, our aim here is to identify a natural assumption we make when modeling data with a conditional generative model, analogously to how unconditional generative models are naturally compatible with data that lives close to a low-dimensional manifold. Continuing the analogy, just like manifolds are used to regularize ill-posed inverse problems, we imagine fiber bundle models can be used as regularizers with additional structure even when the data only approximately lives on a fiber bundle.
} 

Fiber bundles model spaces that are locally product spaces but may have non-trivial topology globally. An example of a space that is globally a product space is the cylinder $\mathcal{E} = S^1 \times [0, 1]$ with the base space being the circle $\mathcal{B} = S^1$ and all fibers being translates of the line segment $\mathcal{Z} = [0, 1]$. The projection $\pi$ associates to each point $x \in \mathcal{E}$ its position along the base circle; $\pi^{-1}$ takes a position along the base circle and returns the corresponding line segment. The same $\mathcal{B}$ and $\mathcal{Z}$ can generate a rather different space---the M\"obius band---which is globally not a product space.\footnote{To explain this mathematically we would need to introduce the transition maps and the fundamental group, which is beyond the scope of our sketch.} 

Thus (under appropriate conditions) the range of a C-Trumpet is locally homeomorphic to a product space (it is a product space up to a stretch), even though it globally need not be depending on the topologies of $\mathcal{Z}$ and $\mathcal{Y}$. The requirement that $\phi$ be a homeomorphism on $U_i$ implies that (locally) the fibers do not intersect. This can of course only hold if $\dim(\mathcal{Y}) + \dim(\mathcal{Z}) \leq \dim(\mathcal{X})$ and the appropriate conditions on the Jacobian are satisfied, but in any case it gives a useful intuition for the kind of problems and datasets that admit modeling by C-Trumpets. 

\begin{figure}
\centering
\begin{subfigure}{.23\textwidth}
  \centering
 \includegraphics[width=0.95\textwidth]{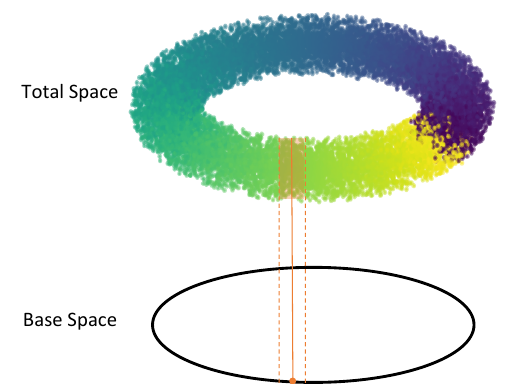}
\caption{Torus}
\label{fig:torus}
\end{subfigure}%
\begin{subfigure}{.23\textwidth}
\centering
\includegraphics[width=0.95\textwidth]{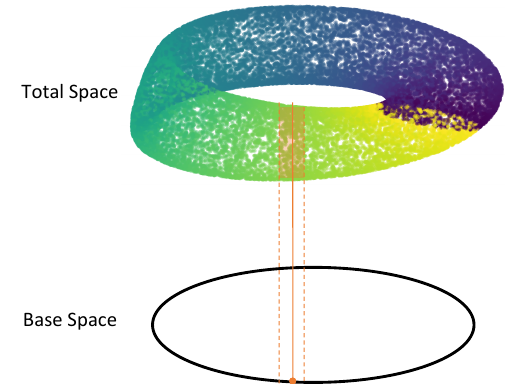}
\caption{Elliptic M\"{o}bius}
\label{fig:mobius}
\end{subfigure}
\caption{Two examples of fiber bundles: Torus and elliptic M\"{o}bius  as  fiber bundles over the base circle.}
\label{fig:fiber bundles}
\end{figure}

\begin{figure}
\centering
\begin{subfigure}{.23\textwidth}
  \centering
 \includegraphics[width=0.82\textwidth]{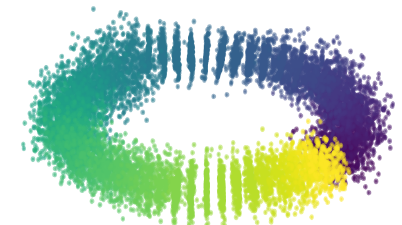}
\caption{Torus}
\label{fig:torus_samples}
\end{subfigure}%
\begin{subfigure}{.23\textwidth}
\centering
\includegraphics[width=0.82\textwidth]{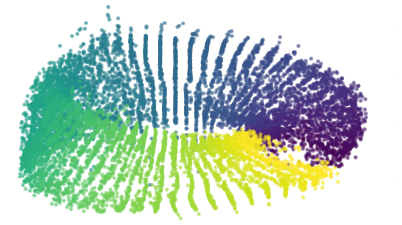}
\caption{Elliptic M\"{o}bius}
\label{fig:mobius_samples}
\end{subfigure}
\caption{Samples on fiber bundles generated by a C-Trumpet; in both (a) and (b) the base manifold $S^1$ is sampled every $6^\circ$.}
\label{fig:fiber bundles generated samples}
\end{figure}

We show that C-Trumpets can indeed model simple low-dimensional fiber bundles: an embedded solid\footnote{By a ``solid'' torus we refer to a bundle whose fibers are disks. Modeling a standard torus is challenging with coupling layers which can only be defined starting in dimension 2. Even in this case their expressivity is limited so this low-dimensional example is only meant as illustration.} torus in Figure~\ref{fig:torus} and a solid M\"{o}bius band with elliptic cross-sections (fibers) in Figure~\ref{fig:mobius}. As shown in Figure~\ref{fig:fiber bundles}, the conditioning samples for both the torus and the elliptic M\"{o}bius are taken to live on the base circle ($y \in [0,2\pi)$). For each angle $y$, C-Trumpets generate samples from a distribution on the disk for the torus and an elliptical disk for the elliptic M\"{o}bius band. Accordingly, we train C-Trumpets with latent dimension two. Figure~\ref{fig:fiber bundles generated samples} demonstrates the generated samples by C-Trumpets; we sample the resulting models in increments of $6^\circ$ to make the fiber bundle structure clear.

\section{Experimental Results}
\label{sec:experimental-results}

We start by showcasing how C-Trumpets provide MMSE, MAP and uncertainty estimates in a variety of imaging inverse problems. The MMSE estimate, $\EX[X|Y=y]$, is approximated by averaging a fixed number of posterior samples from a C-Trumpet fitted to the training data. The MAP estimate can be efficiently approximated using fixed volume-change layers in Section~\ref{sec:fixed_vol_change_layers}, without iterative maximization over $x$. Finally, we compute the uncertainty estimate (UQ) through \rev{a simple pixel-wise standard deviation as}
\begin{equation}
    \begin{array}{rcl}
    \widehat{\text{MMSE}} &=&\displaystyle \dfrac{1}{K} \sum_{k=1}^K f_\theta(z_k;y), \\
    \text{UQ} &=&\displaystyle \sqrt{\dfrac{1}{K} \sum_{k=1}^K |f_\theta(z_k;y) - \widehat{\text{MMSE}}|^2},
    \end{array}
\end{equation}
where $| \cdot |^2$ is applied to each pixel. In all experiments, we use $K=25$ posterior samples to approximate the MMSE and UQ estimates (we find that the quality of MMSE and UQ saturate for a higher number of posterior samples).

We experiment with limited-view computed tomography (CT), nonlinear electromagnetic inverse scattering~\cite{chen2018computational}, and (linearized) seismic travel-time tomography~\cite{kothari2019random}. Additional evaluations on popular compressed-sensing-style benchmarks---denoising, inpainting, super resolution and random masking---are given in Section~\ref{sec:compressed sensing}.

We compare C-Trumpets with two different types of conditional bijective flows, C-INN~\cite{ardizzone2021conditional} and C-Glow~\cite{sorkhei2020full}.  C-INN used conditional coupling layers for the first time, while C-Glow has a rather different architecture by using two parallel bijective flows for simultaneously modeling the target and conditioning samples. In order to emphasize the importance of an expansive injective model over a bijective one, we build a comparison baseline model, C-Rev, which consists of the bijective portion of C-Trumpets with latent-space dimension equaling that of the image data, i.e., $h_\eta (z, y)$. Finally, in Section~\ref{sec: MAP results}, we visually compare the MMSE and MAP estimates for several ill-posed inverse problems.

\subsection{Gaussian Random Fields}
\label{sec:grf}
We choose realizations of Gaussian random fields (GRF) as instances of $p_X(x)$ where we have access to exact posterior distributions. We compare the quality of the posterior approximated by C-Trumpets with the true posterior obtained by analytical solution. The closed-form posterior distribution $p_{X|Y}(x|y)$ where $y = Ax + n$, $x \sim \mathcal{N}(\mu_x, \Sigma_x)$ and $n \sim \mathcal{N}(0, \lambda^2 I)$ as follows,
\begin{equation}
    \begin{array}{cc}
    &p_{X|Y}(x|y) = \mathcal{N}(\mu_{x|y}, \Sigma_{x|y}), \\
    &\mu_{x|y} = \mu_x + \Sigma_{x}A^{T}(A\Sigma_x A^{T} + \lambda^2 I)^{-1}(y - A\mu_x),\\
    &\Sigma_{x|y} = \Sigma_x - \Sigma_x A^{T}(A \Sigma_x A^{T} + \lambda^2 I)^{-1} A \Sigma_x.
    \end{array}
    \label{eq: GRF}
\end{equation}

\begin{figure}
\centering
\includegraphics[width=0.5\textwidth]{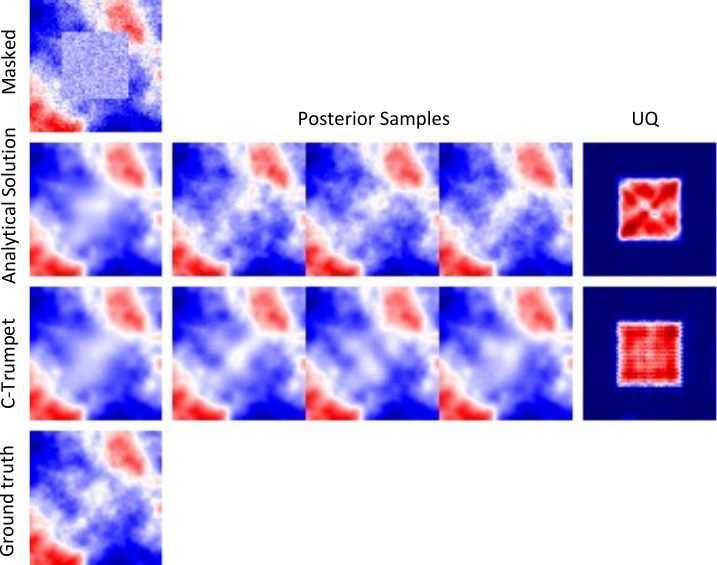}
\caption{Performance comparison between C-Trumpet and analytical solution (true posterior)~\eqref{eq: GRF} in mask problem where the target signals come from a Gaussian distribution; C-Trumpets present MMSE and UQ close to the analytical solution.}
\label{fig: GRF}
\end{figure}

We consider GRFs in resolution $64 \times 64$, $\lambda = 5 \times 10^{-3}$, and let $A$ be the mask operator that replaces a $32 \times 32$ patch at the center of the image with zeros. We train C-Trumpets over $60000$ training samples. In Figure~\ref{fig: GRF} (second to fifth columns), we show the MMSE estimate followed by three posterior samples, and finally the UQ obtained from both the analytical solution~\eqref{eq: GRF} and C-Trumpets. MMSE and UQ are computed over 500 random posterior samples. We compare the MMSE estimates obtained from C-Trumpet and the analytical solution \eqref{eq: GRF} against the ground truth. In Table~\ref{tab:GRF analysis} we find that C-Trumpet performs similar to the analytical solution in both SNR and SSIM metrics. Since the analytical posterior distribution is Gaussian, MAP and MMSE estimators are equal. We find that the MAP and MMSE estimates obtained from C-Trumpet are quite close: an SNR of 44.43dB hinting at the efficacy of our way of computing MAP estimates.

\begin{table}[ht!]
\centering
\renewcommand{\arraystretch}{1}
\caption{Performance comparison of MMSE estimates between C-Trumpets and the analytical solution obtained from~\ref{eq: GRF}; C-Trumpets presents MMSE estimates close to the analytical solution, which shows the effectivity of the proposed method.}
\label{tab:GRF analysis}
    \resizebox{0.35\textwidth}{!}{%
    \begin{tabular}{@{}lccccc@{}}
    \hline
    & SNR (dB) & SSIM \\
    \hline
    C-Trumpets   & 19.71 & 0.89  \\
    Analytical solution & \textbf{20.22} & \textbf{0.91}  \\
    \hline
      \end{tabular}}
\end{table}

\subsection{Computational imaging}
\label{sec:computational imaging}

\begin{figure}
\centering
\begin{subfigure}{.125\textwidth}
  \centering
 \includegraphics[width=0.95\textwidth]{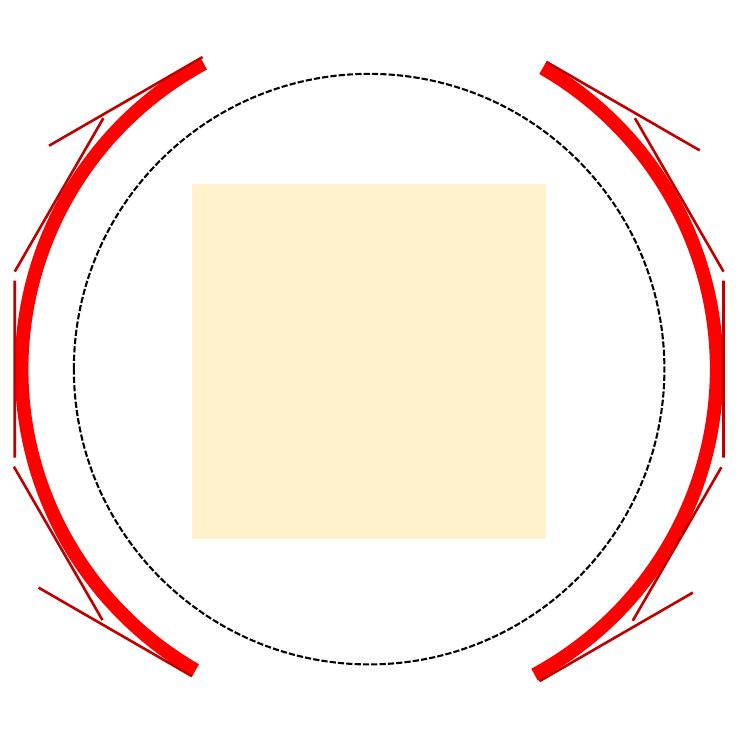}
\caption{CT vertical}
\label{fig:setup CT vertical}
\end{subfigure}%
\begin{subfigure}{.125\textwidth}
\centering
\includegraphics[width=0.95\textwidth]{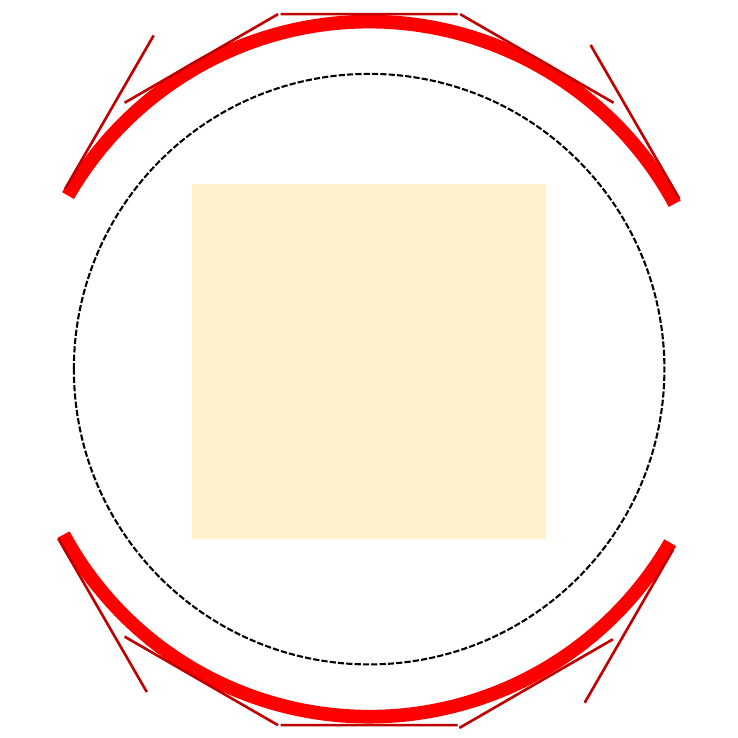}
\caption{CT horizontal}
\label{fig:setup CT horizontal}
\end{subfigure}%
\begin{subfigure}{.125\textwidth}
\centering
\includegraphics[width=0.95\textwidth]{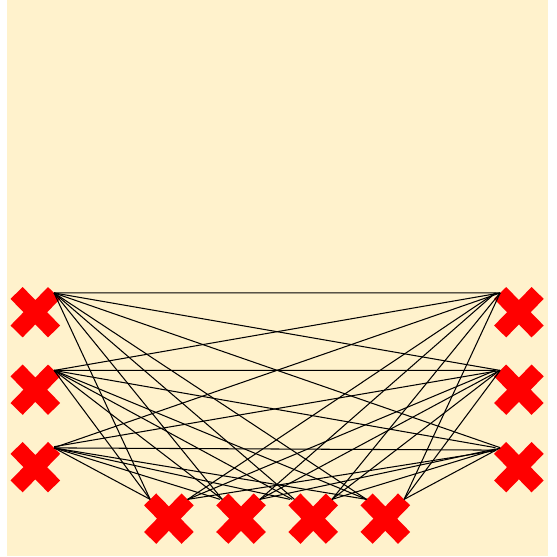}
\caption{Travel time}
\label{fig:setup travel-time}
\end{subfigure}%
\begin{subfigure}{.125\textwidth}
\centering
\includegraphics[width=0.95\textwidth]{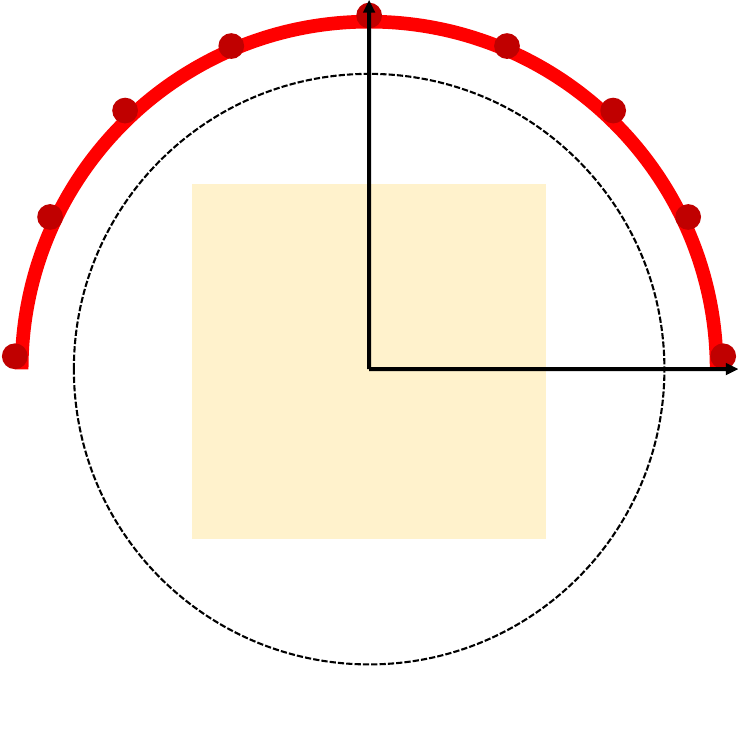}
\caption{Scattering}
\label{fig:setup scattering}
\end{subfigure}
\caption{Sensing geometry for the various imaging problems explored in Section \ref{sec:experimental-results}. The red regions show the locations of the sensors. In the case of CT ((a) and (b)), the line segments indicate that ``sensors'' measure entire projection images as opposed to complex scalars in nonlinear scattering (d).}
\label{fig:sensing geo}
\end{figure}

\begin{figure*}
\centering
\begin{subfigure}{.5\textwidth}
  \centering
 \includegraphics[width=0.9\textwidth]{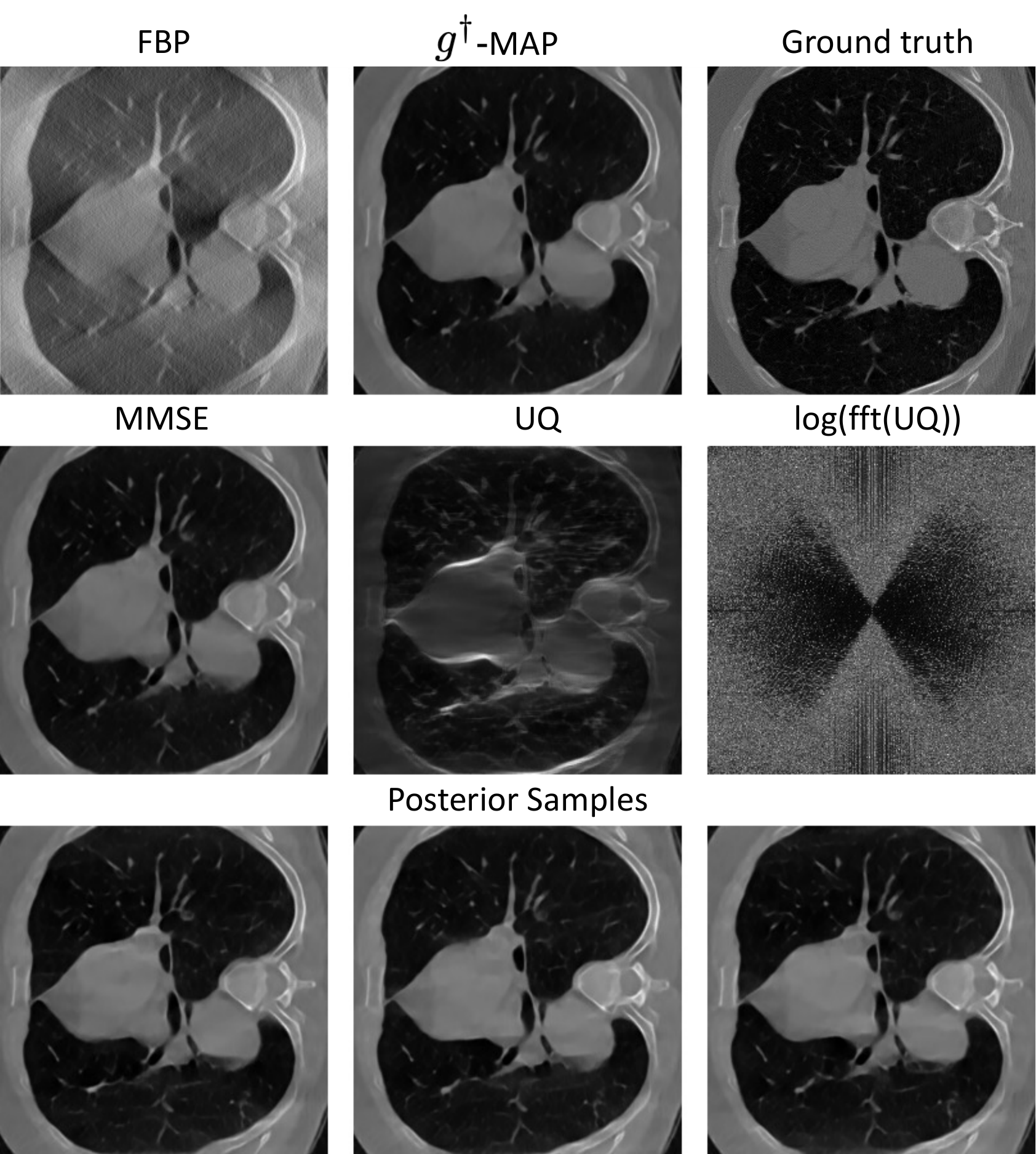}
\caption{Vertical missing cone ($60^\circ$ to $120^\circ$)}
\label{fig:limited-CT_60-120_256}
\end{subfigure}%
\begin{subfigure}{.5\textwidth}
\centering
\includegraphics[width=0.9\textwidth]{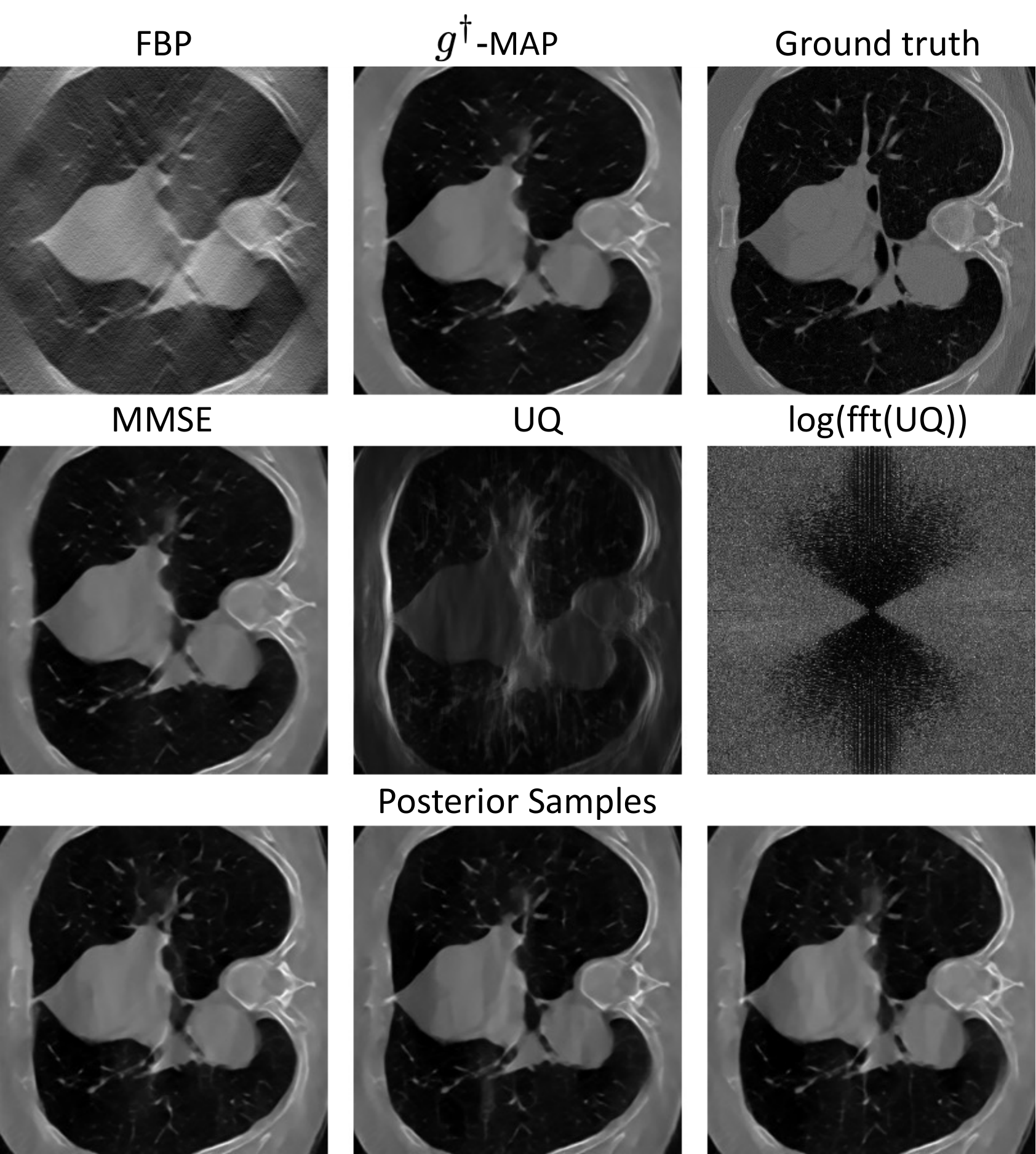}
\caption{Horizontal missing cone ($-30^\circ$ to $30^\circ$)}
\label{fig:limited-CT_0-30_256}
\end{subfigure}
\caption{Limited-view CT in  resolution $256 \times 256$. The frequency-domain uncertainty estimate shows the log-scale standard deviation of DFT-bins. Alignment with the theoretical prediction from the Fourier slice theorem signifies physically-meaningful uncertainty quantification; higher uncertainty (bright regions) inside the missing cone.
}
\label{fig:limited-CT_256}
\end{figure*}

\paragraph{Limited-view CT} CT is a major medical imaging modality. The corresponding inverse problem is to recover an image from its integrals along straight lines, arranged in a so-called sinogram. In limited-view CT (cf. cryo-electron tomography~\cite{doerr2017cryo} and dental CT~\cite{venkatesh2017cone}), a contiguous cone of angles is missing from the acquisition; Figures~\ref{fig:setup CT vertical} and \ref{fig:setup CT horizontal}, illustrate vertical and horizontal missing cones of $60^\circ$. As there are no measurements in a vertical (horizontal) missing cone, we should expect higher uncertainty in horizontal (vertical) components. We use the filtered back projection (FBP) reconstruction as our measurements $y$, and train on 40000 $256 \times 256$ samples from the LoDoPaB-CT ~\cite{leuschner2019lodopab} dataset. The measurement SNR is set to 40 dB. In Figure~\ref{fig:limited-CT_256}, we show posterior samples, {\map}, MMSE, and uncertainty quantification (UQ) estimates from C-Trumpets.
The real-space UQ estimate shows higher uncertainty in the vertical (horizontal) components where we have horizontal (vertical) missing cone. This is consistent with our expectations from the physics of CT.

We further show the UQ estimate in the Fourier domain computed by averaging over the individual DFT bins. Quite pleasingly, this estimate aligns perfectly with the theoretical prediction from the Fourier slice theorem; higher uncertainty (bright regions) are indeed inside the missing cone while it is worth emphasizing that C-Trumpets are not specifically designed for the forward operator of the CT problem. Moreover, as expected, there is higher uncertainty in higher frequency components (see Figure~\ref{fig:Limited-CT_256 samples} in the supplemental materials for more samples). We emphasize that C-Trumpets are the only conditional generative architecture that gives such physically meaningful posterior samples. 

It is also worth mentioning that C-Trumpets model this $256 \times 256$ resolution dataset with only 10M trainable parameters in less than 24 hours of training time on a single NVIDIA V100 GPU. Due to memory constraints, we could simply not train the baseline bijective models C-INN, C-Glow, and C-Rev on $256 \times 256$ images. The large latent space dimension of these models leads to very large memory footprints.

\begin{figure*}
\centering
\begin{subfigure}{.5\textwidth}
  \centering
 \includegraphics[width=0.95\textwidth]{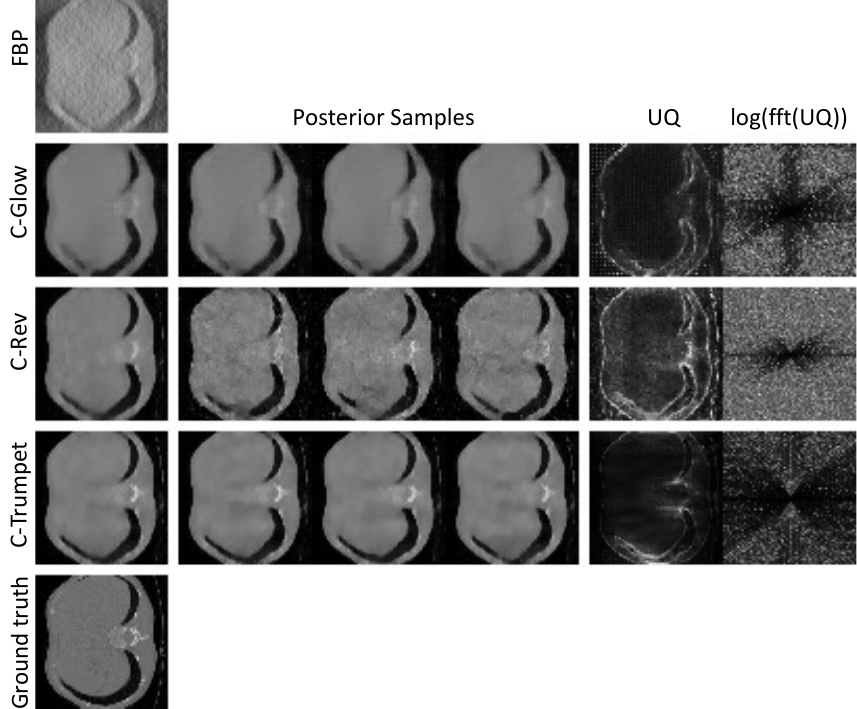}
\caption{Vertical missing cone ($60^\circ$ to $120^\circ$)}
\label{fig:limited-CT_60-120_comparison}
\end{subfigure}%
\begin{subfigure}{.5\textwidth}
\centering
\includegraphics[width=0.95\textwidth]{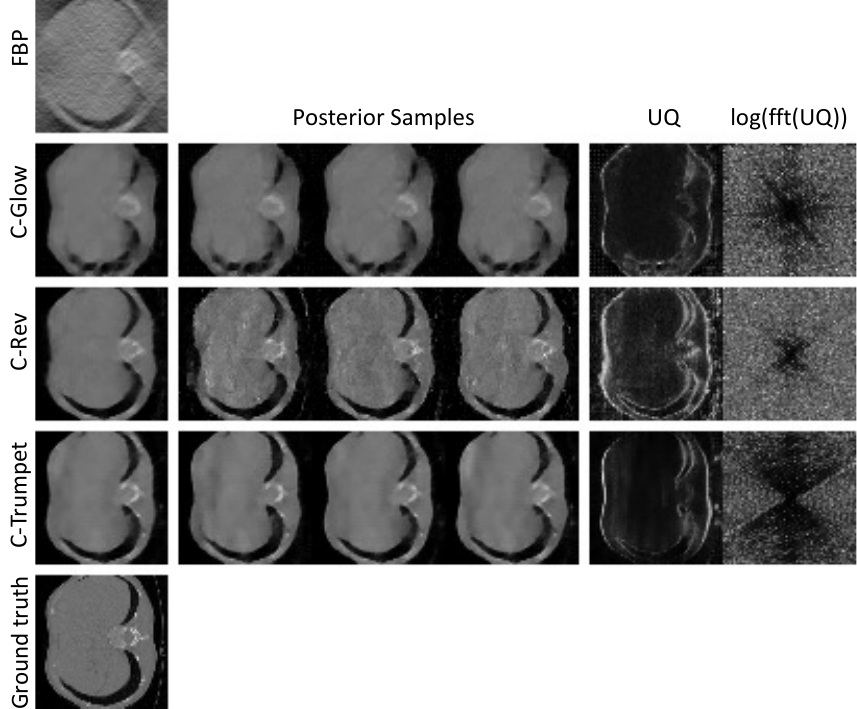}
\caption{Horizontal missing cone ($-30^\circ$ to $30^\circ$)}
\label{fig:limited-CT_0-30_comparison}
\end{subfigure}
\caption{Performance comparison in limited-view CT with resolution $64 \times 64$. The frequency-domain uncertainty estimate shows the log-scale standard deviation of DFT-bins. The estimated $\log(\text{fft}(\text{UQ}))$ by C-Trumpets  aligns with the theoretical prediction from the Fourier slice theorem; higher uncertainty (bright regions) inside the missing cone.}
\label{fig:limited-CT comparison}
\end{figure*}

We can still compare our uncertainty estimates with these models at a lower resolution of $64 \times 64$. Figure~\ref{fig:limited-CT comparison} shows such an experiment at the SNR of 25dB. In the first panel (second to fifth row), we show MMSE estimates of different flow models and three random posterior samples. We further provide UQ estimates in real and Fourier space. As we can see, not only do C-Trumpets outperform the conditional bijective flows in terms of reconstruction quality (MMSE estimate) (see also Table~\ref{tab:quantitative results}) but they also give a more meaningful uncertainty estimate even in high noise. Bijective models generate significant uncertainty outside the missing cone, indicating that they do not learn a reconstruction map consistent with the physics of CT.

\begin{figure*}
\centering
\includegraphics[width=0.7\textwidth]{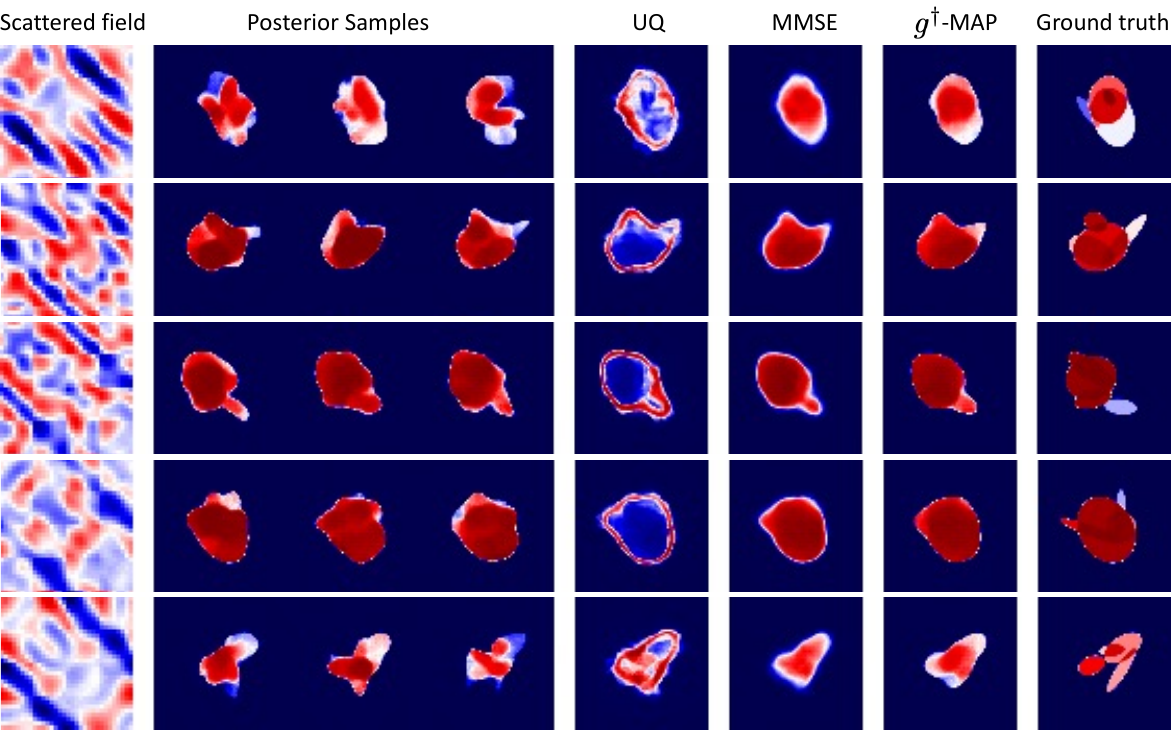}
\caption{Performance of C-Trumpets on nonlinear inverse scattering with $\epsilon_r=6$ with a full-view sensing geometry. Conditioning inputs are scattered fields.}
\label{fig:scattering_samples_full_er6_scattered_conditioned}
\end{figure*}

\paragraph{Electromagnetic inverse scattering} We consider the non-linear electromagnetic inverse scattering problem  as described in~\cite{chen2018computational}.
We consider reconstruction of the finite number of parameters from the scattered fields. Although inverse scattering becomes well-posed and Lipschitz stable with continuous measurements~\cite{nachman1996global}\footnote{This paper considers 2D inverse scattering problem, 3D case is addressed in~\cite{sylvester1987global}},
it is an ill-posed inverse problem with finite number of measurements, which means that noise in the measurements may translate to exponential errors in the reconstruction~\cite{chen2018computational}. Moreover, the problem becomes increasingly non-linear as the relative permittivity of the objects being imaged, $\epsilon_r$, increases. In the first experiment, we use 36 incident plane waves and 36 receivers, distributed uniformly around the object with maximum permittivity of $\epsilon_r = 6$ and dimension 20 cm $\times$ 20 cm. We work at the frequency of 3 GHz. and simulate the measurements by solving the Helmholtz equation explained in Section~\ref{sec: scattering} in the supplemental materials. We add noise to the measurements for a target measurement SNR of 30 dB. We build a dataset of $64 \times 64$ images of overlapping ellipses with 60000 samples.

Figure~\ref{fig:scattering_samples_full_er6_scattered_conditioned} shows the performance of C-Trumpets where the scattered fields are used as conditioning samples. This experiments clearly shows that C-Trumpets can generate meaningful posterior samples, even for a highly non-linear ($\epsilon_r = 6$) and ill-posed problem. Additional experiments with different contrasts $\epsilon_r$ and conditioning schemes (scattered fields vs back projections) are illustrated in Figures~\ref{fig:scattering_samples_er1.5},~\ref{fig:scattering_samples_er2},~\ref{fig:scattering_samples_er6} and~\ref{fig:scattering_samples_slice_er6} in the supplemental materials.

In order to assess the performance of different conditional flows in uncertainty quantification, we design another experiment with 180 incident plane waves and 180 receivers only on the \textit{top} side of the object (\ref{fig:setup scattering}). In this setup, we expect higher uncertainty in the lower part of the object.

\begin{figure}
        \centering
       \includegraphics[width=0.45\textwidth]{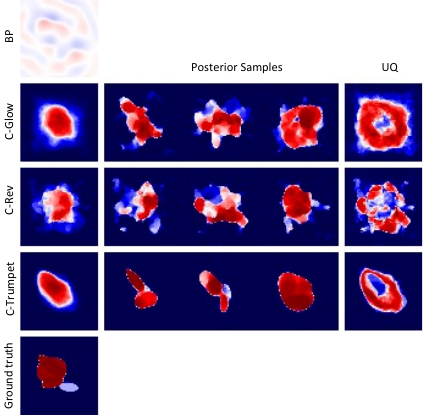}
        \caption{Performance comparison in electromagnetic inverse scattering ($\epsilon_r = 6$, top-view, BP conditioned); C-Trumpets demonstrate more meaningful UQ by assigning higher uncertainty in the bottom of the object (red regions), which lacks measurements.}
        \label{fig:scattering_comparison}
\end{figure}

Figure~\ref{fig:scattering_comparison} illustrates the performance of C-Glow, C-Rev and C-Trumpets for $\epsilon_r = 6$. We use a simple back projection (BP)\cite{belkebir2005superresolution} as conditioning measurements $y$. In the first panel (from the second row to the fifth row), we show MMSE estimates of different flow models: C-Trumpets perform better than bijective flows (see also Table~\ref{tab:quantitative results}). As expected, however, the reconstructions are not as good as in the previous experiment since we only have measurements on one side of the domain. 
We see that C-Trumpets again provide meaningful uncertainty estimates, with more uncertainty in the lower part of the object (red regions in the UQ panel correspond to higher uncertainty).

\begin{figure}
        \centering
       \includegraphics[width=0.45\textwidth]{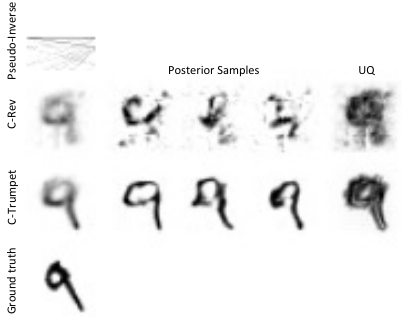}
        \caption{Performance comparison in seismic travel-time tomography with 10 sensors. C-Trumpets demonstrate better posterior samples and MMSE estimate, both models assign higher uncertainty in top of the image (black regions) which lacks measurements.}
       \label{fig:traveltime_comparison}
\end{figure}

\paragraph{Seismic travel-time tomography (NS)} We work with the linearized seismic tomography operator as described in \cite{kothari2019random}. Here we are given travel times of a seismic wave between each sensor pair in a network of NS sensors placed on the ground. The travel times are assumed to linearly depend on the ``slowness'' map which is taken to be an MNIST image~\cite{lecun1998gradient}. We use NS$=10$ sensors on the boundary of the lower part of the domain which yields 33 measurements as shown in Figure \ref{fig:setup travel-time}. We use the pseudo-inverse of the measurements $y$ as the conditioning samples and work at the SNR of 40dB. Figure~\ref{fig:traveltime_comparison} compares the performance of C-Trumpets and C-Rev. In the first panel, the second and the third rows show MMSE estimates of C-Rev and C-Trumpets. C-Trumpets outperform C-Rev in both MMSE estimate and posterior sampling (see also Table~\ref{tab:quantitative results}). Furthermore, given no sensors (Figure~\ref{fig:setup travel-time}) in the top regions of the image (or slowness map), we would expect higher uncertainty there. The UQ column shows that estimates from C-Trumpets assign higher uncertainty to the top half of the domain compared to C-Rev. Additional results are shown in Figure~\ref{fig:traveltime_4} in the supplemental materials.


\subsection{Image restoration}
\label{sec:compressed sensing}
In this section, we compare C-Trumpets with models on various image restoration tasks. Similarly as in computational imaging problems, each task requires training a different model but once trained, the model can be used instantaneously for arbitrary measurements $y$. We consider four standard restoration tasks: (i) \textbf{Image denoising}: We train a model to generate plausible clean images given a noisy image at an SNR of $-1$ dB; (ii) \textbf{Image super resolution} ($f$): Generate high-resolution images given an image downsampled by a factor of $f$ along each axis; (iii) \textbf{Random mask} ($p$): The degradation process replaces every pixel with zero with probability $p$; and (iv) \textbf{Mask} ($s$): The degradation process replaces an $s \times s$ patch of the image with zeros.

We used the 8-bit RGB CelebA~\cite{liu2015deep} dataset with 80000 $64 \times 64$ training samples. Figure~\ref{fig: Compressed sensing comparison} in the supplemental materials compares the performance of flow models on different image restoration tasks.  Table~\ref{tab:quantitative results} further gives the SNR and SSIM of the MMSE estimate. C-Trumpets consistently outperform other conditional bijective flows. The difference in the performance of C-Rev and C-Trumpets suggests that the low-dimensional latent space of C-Trumpets acts as an effective regularizer in the restoration mapping, $f_\theta$. C-Trumpets also provide a meaningful uncertainty quantification. For example, although the forward operator is random in the random mask problem, C-Trumpets still capture a meaningful uncertainty estimate by assigning higher uncertainty inside the masked region (see Figure~\ref{fig:mask32_compare}).

In order to assess the memory requirements of the different models, we compare the number of trainable parameters used for training over $64 \times 64$ RGB images: C-INN: 13M, C-Glow: 14M, C-Rev: 22M and C-Trumpets: \textbf{4M}. We did not have sufficient resources to train bijective models over $256 \times 256$ images but it is clear that the differences in the memory footprint at that resolution would be further exacerbated.

\begin{table}
\renewcommand{\arraystretch}{1.3}
\caption{Performance of MMSE estimate (computed over 25 posterior samples) of different models on solving inverse problems averaged over 5 test images}
\label{tab:quantitative results}
\begin{subtable}{.49\textwidth}
      \centering
    \caption{SNR (dB)}
    \label{tab:compressed_sensing_SNR}
    \resizebox{0.95\textwidth}{!}{%
    \begin{tabular}{@{}lccccc@{}}
    \hline
    & C-INN & C-Glow & C-Rev & C-Trumpets \\
    \hline
    {\textit{Denoising }}   & 15.98 & 15.67 &  15.99 & \textbf{16.86}  \\
    {\textit{Super resolution }($\times 4$)} & 18.57 & 19.24 &  19.22  & \textbf{20.70} \\
    {\textit{RandMask }($p=0.2$)} & \textbf{20.84} & 19.70  & 13.31  & 20.11   \\ 
    {\textit{Mask} ($s=32$)}    & 18.91 &  17.86 & 18.54  &  \textbf{21.01} \\
    {\textit{Limited-view CT} }  & - & 11.63 &  13.13  &  \textbf{13.58} \\
    {\textit{Scattering ($\epsilon_r = 6$)} } & - & 1.45 & 1.19  & \textbf{4.04}\\
    {\textit{Travel-time (NS = 10)} } & - & - &  14.83  &  \textbf{18.19} \\
    \hline
    \end{tabular}}
\end{subtable}

\bigskip
\begin{subtable}{.49\textwidth}
      \centering
    \caption{SSIM}
    \label{tab:compressed_sensing_SSIM}
    \resizebox{0.95\textwidth}{!}{%
    \begin{tabular}{@{}lccccc@{}}
    \hline
    & C-INN & C-Glow & C-Rev & C-Trumpets \\
    \hline
    {\textit{Denoising}} & 0.63 & 0.60 &  0.66 & \textbf{0.70}  \\
    {\textit{Super Resolution }($\times 4$)} & 0.79  & 0.82  & 0.79  & \textbf{0.84} \\
    {\textit{RandMask }($p=0.2$)} & 0.86 & 0.82 & 0.57  & \textbf{0.87}   \\ 
    {\textit{Mask} ($s=32$)}&  0.83 & 0.80 &  0.84  &  \textbf{0.88} \\
    {\textit{Limited-view CT} }  & - & 0.59 &  0.71  &  \textbf{0.74} \\
    {\textit{Scattering ($\epsilon_r = 6$)} } & - & 0.67 &  0.65  &  \textbf{0.73} \\
    {\textit{Travel-time (NS = 10)} } & - & - &  0.59  &  \textbf{0.64} \\
    \hline
    \end{tabular}}
\end{subtable} 
\end{table}

\subsection{MAP vs MMSE for ill-posed inverse problems}
\label{sec: MAP results}
\begin{figure}
\centering
\includegraphics[width=0.48\textwidth]{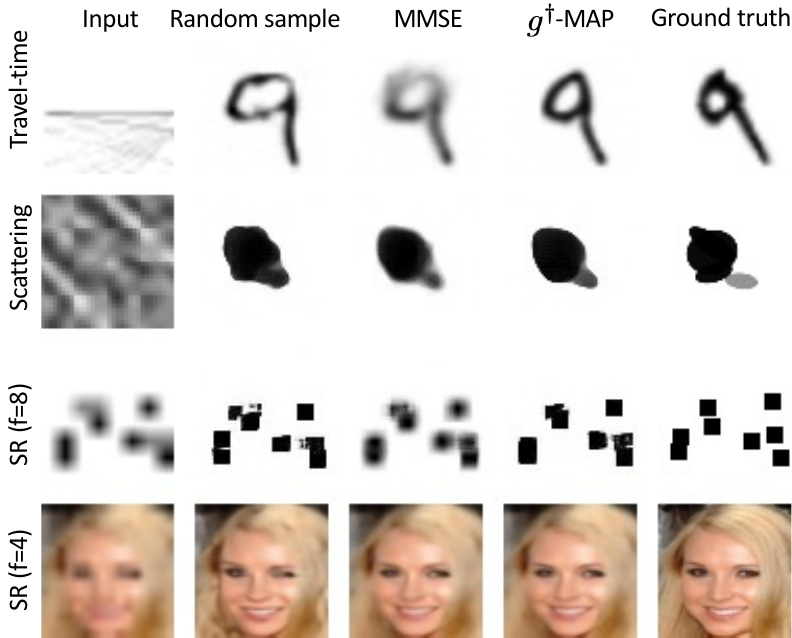}
\caption{MMSE and {\map} estimations in different inverse problems; the proposed {\map} estimate gives much sharper reconstruction than MMSE.}
\label{fig: MAP_MMSE_samples}
\end{figure}

Figure~\ref{fig: MAP_MMSE_samples} demonstrates the MMSE, {\map} and a random posterior sample for four types of ill-posed inverse problems. The MMSE estimate is obtained by averaging over 25 posterior samples.  Although the MMSE estimate is the optimal reconstruction in terms of the $\ell^2$-error, we see from Figure~\ref{fig: MAP_MMSE_samples} that it is often blurry, especially when the true posterior is multi-modal; {\map} estimates are sharper. Moreover, as the MMSE estimate is obtained by averaging over the posterior samples, it is not generally on the manifold, while the {\map} estimate is always on the manifold.

\section{Related Work}
There is by now a very large body of work on solving inverse imaging problems using deep neural networks. On the supervised regression end of the spectrum arguably the most important architecture is the U-Net~\cite{ronneberger2015u}. It has been applied with great success to a variety of imaging problems including CT~\cite{jin2017deep}, magnetic resonance imaging (MRI)~\cite{hyun2018deep} and electromagnetic inverse scattering~\cite{wei2018deep}. Its success may be attributed to the particular multiscale structure~\cite{lee2017deep,liu2020learning} which matches both the physical description of the imaging problems and the representation of the involved image classes. Many alternatives have been proposed for specific problems where the assumptions that make the U-Net a natural choice do not hold, for example for wave-based problems~\cite{fan2019solving,kothari2020learning}.

On the other hand, trained generative models have been shown to be effective priors~\cite{bora2017compressed,pmlr-v119-asim20a} in ill-posed inverse problems that can be trained in an unsupervised manner. Normalizing flows in particular were used to approximate MAP estimates using iterative optimization \cite{pmlr-v119-asim20a,whang2020compressed}. However, normalizing flows are bijective, requiring large memory and compute budget even at moderate resolution. Moreover, they lack a  low-dimensional latent space which has been shown to effectively regularize the inversion. Brehmer and Cranmer \cite{brehmer2020flows} proposed the first injective model for densities supported on low-dimensional manifolds. Kothari et al. \cite{kothari2021trumpets} proposed injective flows with significantly optimized compute and memory requirements which were shown to outperform earlier variants of injective and bijective flows in computational imaging problems.

A variety of methods have been proposed for Bayesian imaging. The latter aims to approximate posterior distribution and / or the various point estimators related to the posterior. The authors of~\cite{repetti2019scalable} consider a \textit{convex} log-likelihood function for posterior distribution around the MAP estimate, which is suitable for modeling unimodal posteriors. Normalizing flows have been proposed as variational approximators to the posterior distribution for a given measurement~\cite{sun2020deep}. The authors of~\cite{whang2021composing,kothari2021trumpets} propose to train a flow model to approximate the posterior corresponding to a prior which is also modeled using a flow model. More recently, pre-trained GANs were used in conjunction with MCMC to generate posterior samples in non-linear inverse problems~\cite{bohra2022bayesian}. The authors of ~\cite{bhadra2022mining,marinescu2020bayesian} use a style-GAN generator~\cite{karras2019style} to regularize ill-posed inverse problems and generate posterior samples. 
All these methods train a new generative model or run an iterative process for every measurement, which makes them slow when applied to multiple reconstructions. Further, training for each conditioning sample requires many calls to the forward operator. To tackle these issues, one may consider \textit{amortized inference} to make the generative models conditioned based on the measurements.

Conditional versions of generative adversarial networks (GANs)~\cite{mirza2014conditional, goodfellow2014generative} and variational autoencoders (VAEs)~\cite{sohn2015learning, kingma2013auto} rely on injecting conditioning data into the different layers of the generator model.
However, the lack of access to the posterior distribution make conditional GANs difficult to be used in inference tasks.
\rev{On the other hand, VAEs provide lower bounds on likelihoods of generated samples. While these bounds can be made tighter by importance weighting~\cite{burda2015importance}, C-Trumpets allow one to compute exact end-to-end likelihoods via stochastic estimation, as well as to obtain fast exact values of likelihoods before the high-dimensional expansion.}
These generative models also suffer from mode collapse and training instabilities. 
Conditional normalizing flows were introduced in \cite{ardizzone2021conditional} to estimate the posterior by modifying the scale and shift terms of the coupling layers. The authors of \cite{winkler2019learning} additionally make the mean and covariance  of the base Gaussian distribution depend on the measurements. More recently, \cite{lu2020structured} proposed to append the measurements to all layers of the Glow network~\cite{kingma2018glow} in order to enable greater information flow from conditioning data to the generated samples.
A different approach to conditioning the flow models has been developed by~\cite{sorkhei2020full,pumarola2020c}, benefitting from two parallel flow models for simultaneously modeling of target and conditioning samples. All these conditional normalizing flows are bijective mappings from latent space to the target domain for each measurement. It is worth mentioning that all these models can be used in the bijective part of C-Trumpets to exploit their advantages in posterior modeling.
On the side of theory, Puthawala et al. \cite{puthawala2021universal} establish universality of density and manifold approximation of injective flows such as those in \cite{brehmer2020flows, kothari2021trumpets}. 

\section{Limitations and Conclusions} \label{sec:discussion}

We proposed C-Trumpets, a conditional injective flow model that enables amortized inference with approximate posteriors that live on low-dimensional manifolds. Our proposed model is considerably cheaper to train in terms of memory and compute costs compared to the regular conditional flows. The experiments we performed indicate that C-Trumpets generate better posterior samples and more accurate uncertainty estimates over a variety of ill-posed inverse problems. The proposed fixed-volume-change coupling layers enable us to approximate the sharp MAP estimates instantaneously after training. High computational demands of training bijective flows at high resolution have thus far impeded their wider adoption in computational imaging workflows. The comparably lightweight memory footprint of C-Trumpets together with physically-consistent UQ makes them an attractive architecture for imaging problems where characterizing uncertainty is paramount.

\paragraph*{Limitations} C-Trumpets have several limitations that warrant discussion. The latent space dimension in C-Trumpets is chosen arbitrarily and it may be quite different from the true dimension of the posterior support. Recent work~\cite{zhang2021flow} proposes an injective flow architecture that estimates the dimension of the data manifold. Similar ideas may extend to C-Trumpets but for the moment we rely on rules of thumb rather than principled choices. Another limitation is that it is not straightforward to estimate the likelihoods of samples generated by C-Trumpets (cf. Section~\ref{sec:fixed_vol_change_layers}), the reason being that the Jacobian determinant of compositions of maps between spaces of different dimension cannot be written as a product of Jacobian determinants of the constituent maps. Likelihood estimates can still be obtained by sampling but that is considerably slower than what is possible with bijective flows. Recently, Ross et al.~\cite{ross2021tractable} proposed an injective generator which provides access to the exact likelihood of the generated samples, but the constraints they impose on the architecture in order to enable this feat seem to severely limit expressivity. The design of an injective model that is at once expressive, lightweight, and gives fast exact likelihoods remains an open problem. On the theoretical side, further studies are needed to characterize the types of posterior distributions that can be modeled by C-Trumpets, especially with fixed-volume-change layers. The important open question is that of universality of C-Trumpets as models of conditional distributions. \secrev{Finally, Siahkoohi et al.~\cite{siahkoohi2021preconditioned, siahkoohi2022reliable} fully exploit the depth-independent memory complexity of normalizing flows to handle high-dimensional data. This strategy can also be used in C-Trumpets to further improve memory efficiency and apply the model to super high-dimensional imaging problems.}

\bibliographystyle{IEEEtran}
\bibliography{main}


\vfill
\pagebreak
\begin{center}
\textbf{\large Supplemental Materials: Conditional Injective Flows for Bayesian Imaging}
\end{center}
\setcounter{equation}{0}
\setcounter{figure}{0}
\setcounter{table}{0}
\setcounter{page}{1}
\makeatletter
\renewcommand{\theequation}{S\arabic{equation}}
\renewcommand{\thefigure}{S\arabic{figure}}
\setcounter{section}{0}
\renewcommand{\thesection}{S-\Roman{section}}

\section{Fast Inverses}
\label{sec: faster MSE}

Inverting the revnet coupling layers requires the matrix inverse of the kernel of the $1 \times 1$ convolution layer. As this may be slow, Kingma and Dhariwal~\cite{kingma2018glow} proposed to use the LU decomposition to reduce the computational complexity,
\begin{equation}
    w = PL(U + \text{diag}(s)),
\end{equation}
where $P$ is a permutation matrix, $L$ is a lower-triangular matrix with ones on diagonal and $s$ is a vector. Computing the $\log \det$ of the layer Jacobian then simplifies to
\begin{equation}
    \log |\det(w)| = \sum_{i=1}^d \log(|s_i|).
\end{equation}
The LU decomposition thus reduces the complexity of computing the $\log \det$ of the Jacobian of the $1\times 1$ convolution layer from $\mathcal{O}(c^3)$ to $\mathcal{O}(c)$ (and one-time factorization cost). This trick, however, was only used in inference (when sampling from the model), but not for training the bijective flows as matrix inversion is not critical in that case.
Unlike bijective flows, C-Trumpets do require matrix inversion during the MSE training phase (cf. ~\eqref{eq:lmse_conditional}). In order to reduce the computational cost of inversion, we leverage the LU decomposition as follows
\begin{equation}
    w^{-1} = (U + \text{diag}(s))^{-1}L^{-1}P^{-1}.
\end{equation}
Since $L$ and $U$ are triangular matrices and $P$ is a fixed rotation, computing the inverses costs $\mathcal{O}(2 \times c)$ instead of $\mathcal{O}(c^3)$. This significantly reduces the training time of the injective part of C-Trumpets, especially in high-dimensional problems.

\section{Additional Experimental Details}
\subsection{Fiber bundles}
\noindent \emph{Datasets:} We analyzed the performance of C-Trumpets on two fiber bundle datasets:
\begin{itemize}
    \item Torus
    \begin{equation}
    \begin{array}{l}
         x = \cos(t) [R + r \cos(s)] \\
         y = \sin(t) [R + r \cos(s)] \\ 
         z = r \sin(s)
    \end{array}
    \end{equation}
    where $t \in [0,2\pi)$ and $s \in [0,2\pi)$. We set $R = 1$ and $r = 0.25$ and generate $18000$ training samples.
    
    \item Elliptic M\"{o}bius
    \begin{equation}
    \begin{array}{l}
         x = \cos(t) [R - b \sin(t/2)\sin(s) + a \cos(t/2) cos(s)] \\
         y = \sin(t) [R - b \sin(t/2)\sin(s) + a \cos(t/2) cos(s)] \\ 
         z = b \cos(t/2) \sin(s) + a \sin(t/2) \cos(s)
    \end{array}
    \end{equation}
    where $t \in [0,2\pi)$ and $s \in [0,2\pi)$. We set $R = 1$, $a= 0.4$ and $b = 0.1$ and generate 18000 training samples.
\end{itemize}

\noindent \emph{Network architecture and training details:}
The injective part of C-Trumpets maps $\R^2 \to \R^3$ and consists of an expansion layer (a matrix of size $\R^{3 \times 2}$) followed by 24 revent blocks without activation normalization layer. The bijective part which maps $\R^2 \to \R^2$ also consists of 32 revent blocks. We use coupling layers proposed in~\cite{kingma2018glow} with 3 fully-connected layers for scale and bias. We train for 100 epochs using the Adam optimizer~\cite{kingma2014adam} with a learning rate of $10^{-3}$ for both the injective and the bijective parts of C-Trumpets. We note that this minimal input dimension of the coupling layers has yields poor expressivity, but we only use this example for illustration.

\subsection{Limited-view CT}
\label{app: limited-view CT}
We consider the 2D parallel-beam CT problem with a missing cone of sensors.

\noindent \emph{Network architecture:}
We describe the architecture of C-Trumpets for $256 \times 256$ images.
The injective part consists of 6 injective revnet blocks, each increasing the dimension by a factor of 2. Between the injective layers, we intersperse 36 bijective revnet blocks. We choose a latent space of size 2048. The bijective part  consists of 48 bijective revnet blocks. We use 3 convolutional layers for the scale and bias networks of fixed-volume-change coupling layers. The conditioning networks have 3 conv layers along with a `squeeze' layer that helps match the dimension of the conditioning sample to the input of the scale network.

The architecture of C-Trumpets for $64 \times 64$ images is similar to the $256 \times 256$ variant except that we use 18 (resp. 24) revnet blocks in injective (resp. bijective) subnetworks and choose a 64-dimensional latent space. The C-Rev model used for comparison has 24 revnet blocks, all at the highest resolution (i.e $64 \times 64$).

We initialize the weights of the revnet blocks as in \cite{kingma2018glow}. All elements of the skip connection matrix $S$  are initialized to $0.5$. All models are trained for 300 epochs (150 per phase) using the Adam optimizer~\cite{kingma2014adam} with a learning rate of $10^{-4}$.

\subsection{Electromagnetic inverse scattering}
\label{sec: scattering}

We consider the 2D transverse magnetic inverse scattering problem. We illuminate the domain of investigation $D_{inv}$---a $D{\times} D$ square---by $N_{i}$ plane waves; $N_{r}$ antennas measure the scattered field for each incident wave.

The forward scattering problem can be described by two coupled equations \cite{chen2018computational}. The first equation, called the state equation, relates the total electric field ($E^{t}$) in an unknown domain to its contrast current density ($J$) as
\begin{equation}
    E^{t}(r) = E^{i}(r)+k_{0}^{2}\int\displaylimits_{D_{inv}}g(r,r')J(r')dr',
    \label{State_Equation_IntegralForm}
\end{equation}
where  $g(r,r')=(1/4i)H^{(2)}_{0}(k_{0}|r-r'|)$, $H^{(2)}_{0}$ (a Hankel function of the second kind) is the 2-D free space Green's function, and $J(r')$ can be calculated using the equivalence theorem as $J(r')=\chi(r')E^{t}(r')$ with the contrast $\chi(r')=\epsilon_{r}(r')-1$.

The second equation, called the data equation, maps the contrast current source to the scattered electric field ($E^{s}$) at antenna locations. For $r \in S$,
\begin{equation}
    E^{s}(r) = k_{0}^{2}\int\displaylimits_{D_{inv}}g(r,r')J(r')dr'.
    \label{Data_Equation_IntegralForm}
\end{equation}
The task of inverse scattering is to reconstruct the contrast $\chi$ from the scattered field $E^{s}$.

\noindent \emph{Network architecture:} We use the same architecture and training parameters as in the  $64 \times 64$ 2D-CT problem.

\subsection{Linearized travel-time tomography}
Linearized travel-time tomography is inspired by seismic travel-time tomography that aims to reconstruct the wavespeed variation inside a planet. Sensors are placed on the ground and we measure arrival times of surface waves. We assume that the receivers and transmitting sensors are co-located. We further assume that the pixel intensities represent the ``slowness'' (inverse wave speed). We can therefore measure the travel times between a transmitting sensor, $s_i$ and a receiving sensor $s_j$ as
\begin{equation}
\label{eq:travel_time_eqn}
t(s_i, s_j) = \int_0^1 f(s_i + \lambda (s_j - s_i)) d\lambda,
\end{equation}
where $f$ represents the image. Note that $t(s_i, s_j) = t(s_j, s_i)$.

\noindent \emph{Network architecture:} The injective subnetwork consists of
4 injective revnet blocks, each increasing the dimension by a factor of 2. We use 12 bijective revnet blocks interspersed in between the injective layers. The bijective part takes 64-dimensional latent vectors and consists of 12 bijective revnet blocks. C-Rev used in comparison has 16 revnet blocks all at the $32 \times 32$ resolution. The training hyperparameters are same as the CT problem.

\subsection{Image restoration tasks}
We use the same architecture and training hyperparameters as that of the CT problem except that we work with $64 \times 64 \times 3$ resolution images. The latent dimension of C-Trumpets is chosen to be 192 ($64\times3$).

\section{Additional Experiments}
\label{app: additional experiments}
\subsection{Class-based image generation}
\label{app: class-based image generaation}
\begin{figure*}
\centering
\begin{subfigure}{.5\textwidth}
  \centering
\includegraphics[width=0.95\textwidth]{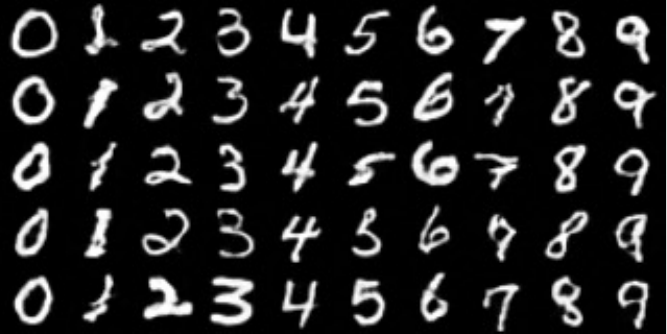}
\caption{MNIST}
\label{fig:MNIST-cs}
\end{subfigure}%
\begin{subfigure}{.5\textwidth}
\centering
\includegraphics[width=0.95\textwidth]{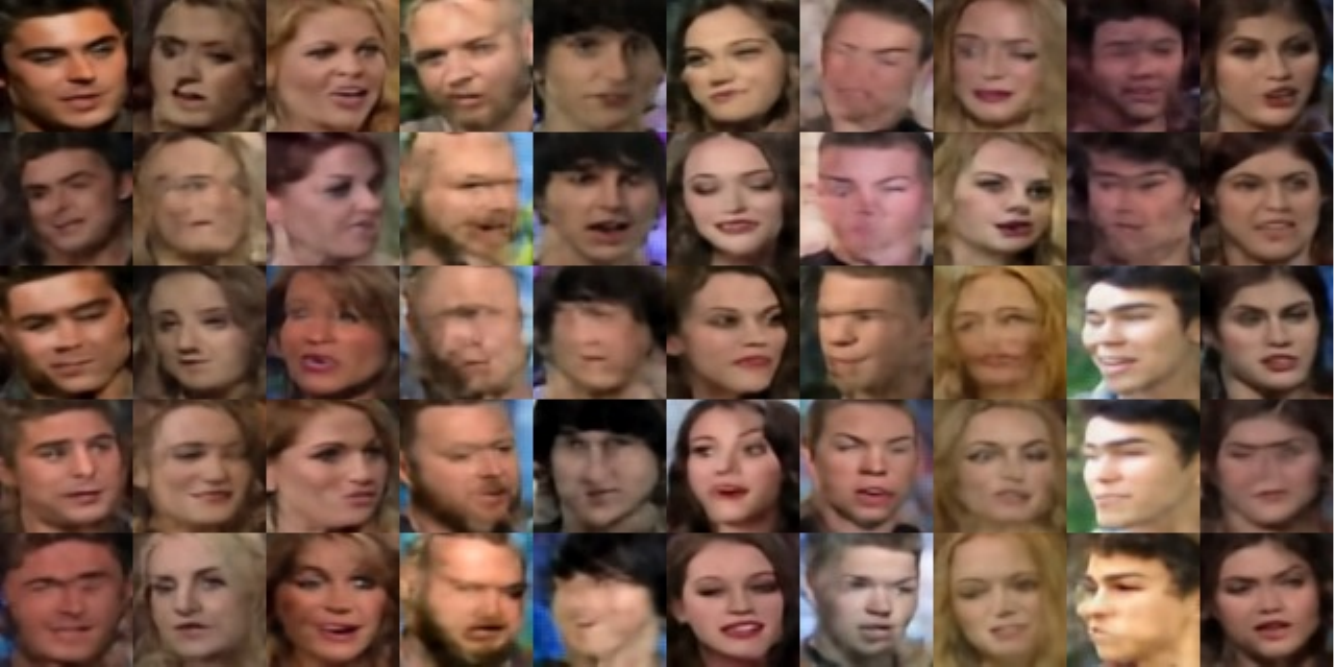}
\caption{Voxceleb}
\label{fig:voxceleb-cs}
\end{subfigure}
\caption{Class-based image generation}
\label{fig: class-based image generation}
\end{figure*}

We perform class-based image generation over MNIST digits~\cite{lecun1998gradient} and a subset of 10 people from the voxceleb~\cite{nagrani2017voxceleb} face dataset with 5000 $64 \times 64$ training samples. We use the one-hot class labels as conditioning vectors to generate samples from the given class and use two fully-connected layers followed by a reshaping module for the conditioning network of C-Trumpets. Figure~\ref{fig: class-based image generation} shows the class-based generated samples by C-Trumpets. This experiment indicates that our proposed model can generate good quality class-conditioned samples.

\subsection{Image restoration and inverse problems}
\label{app: inverse problems}
\begin{figure*}
\centering
\begin{subfigure}{.5\textwidth}
\centering
\includegraphics[width=0.95\textwidth]{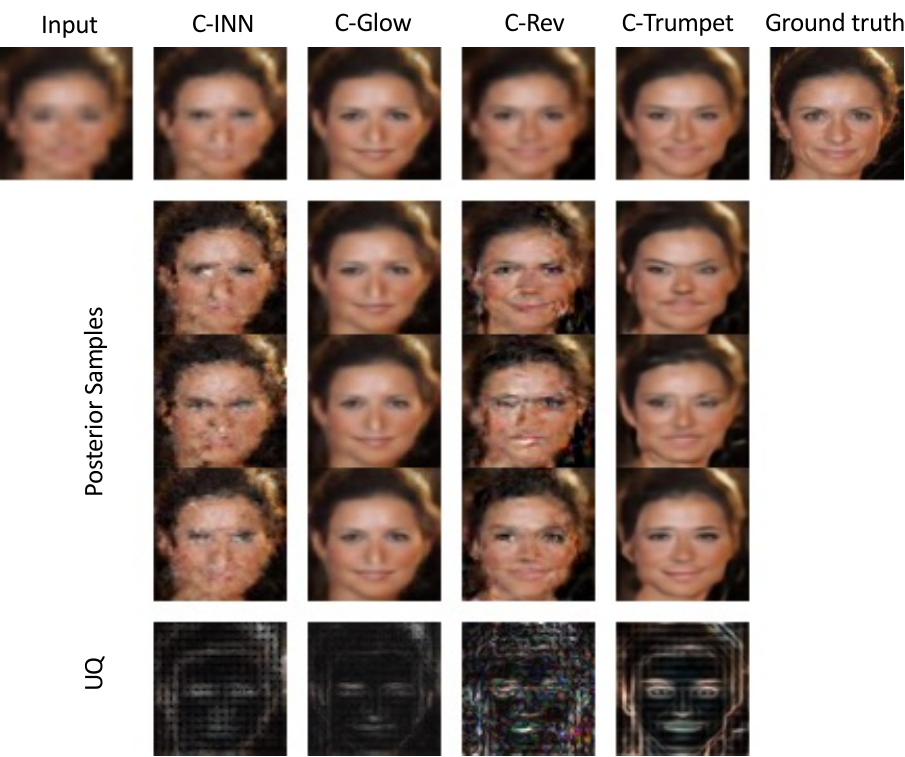}
\caption{Super resolution ($f=4$)}
\label{fig:sr4_compare}
\end{subfigure}%
\begin{subfigure}{.5\textwidth}
\centering
\includegraphics[width=0.95\textwidth]{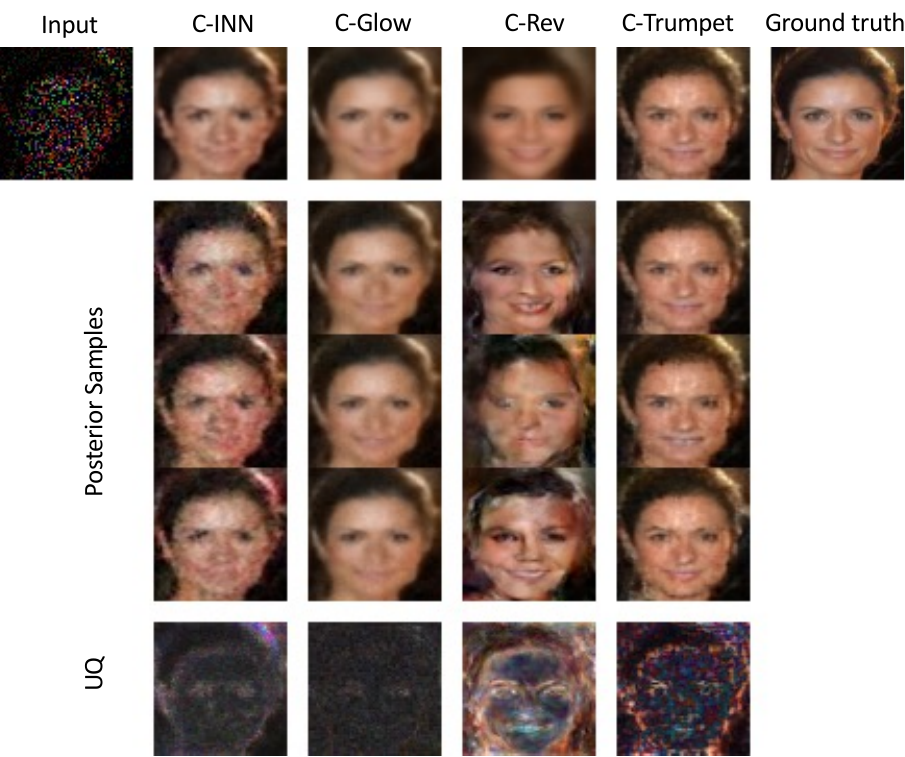}
\caption{Random mask ($p=0.2$)}
\label{fig:random_mask_compare}
\end{subfigure}
\begin{subfigure}{.5\textwidth}
\centering
\includegraphics[width=0.95\textwidth]{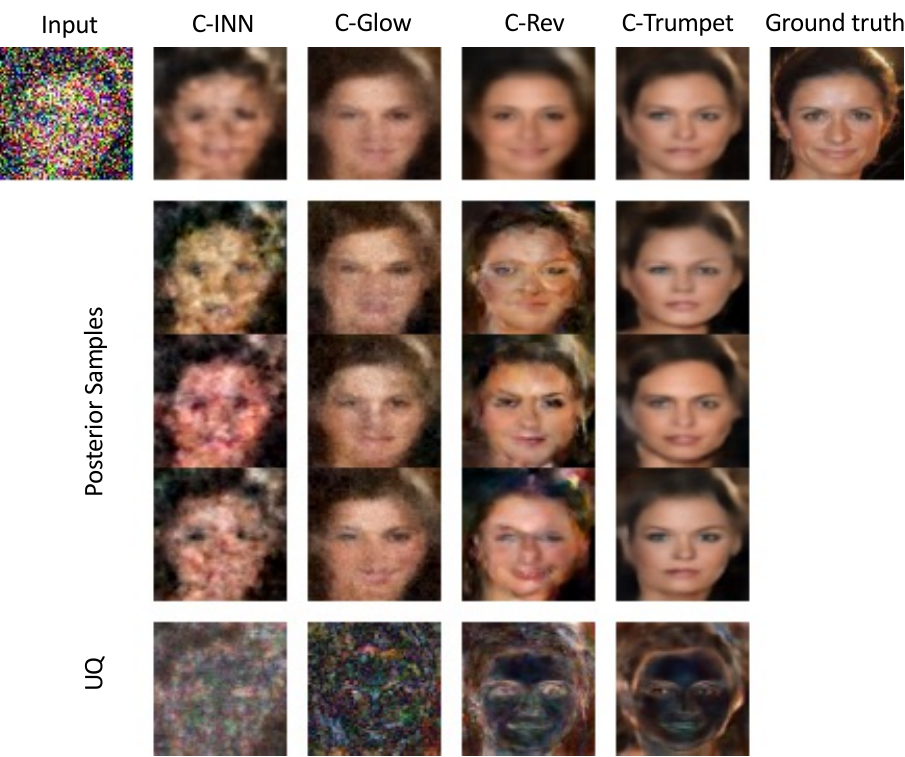}
\caption{Denoising}
\label{fig:denoising_m1_compare}
\end{subfigure}%
\begin{subfigure}{.5\textwidth}
\centering
\includegraphics[width=0.95\textwidth]{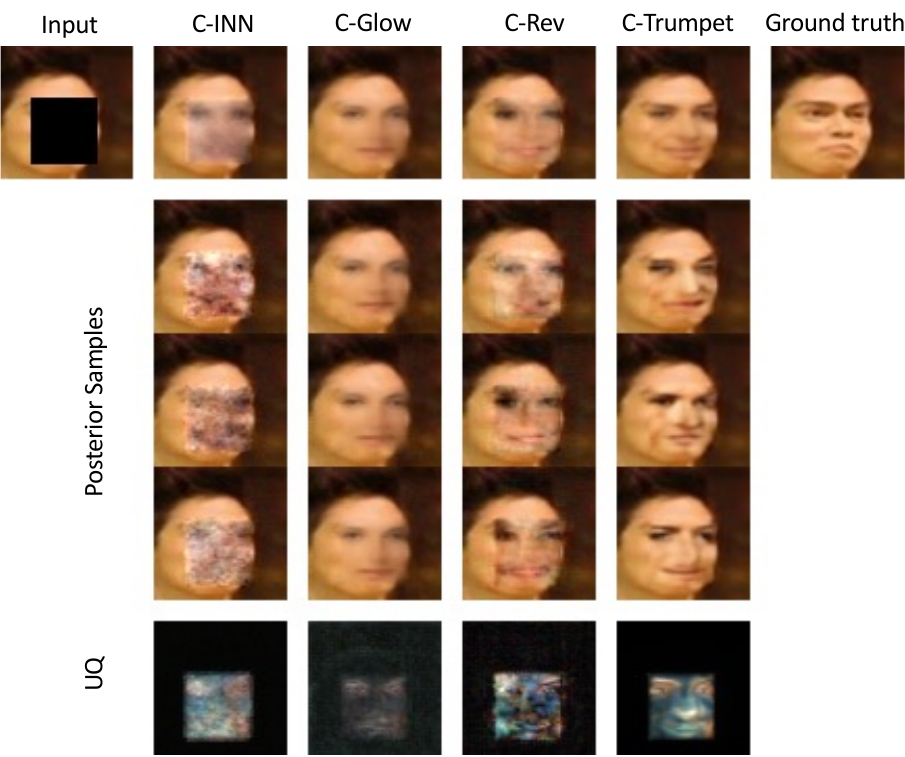}
\caption{Mask ($s=32$)}
\label{fig:mask32_compare}
\end{subfigure}
\caption{Performance comparison over image restoration problems.}
\label{fig: Compressed sensing comparison}
\end{figure*}

\begin{figure*}
\centering
\begin{subfigure}{\textwidth}
\centering
\includegraphics[width=0.9\textwidth]{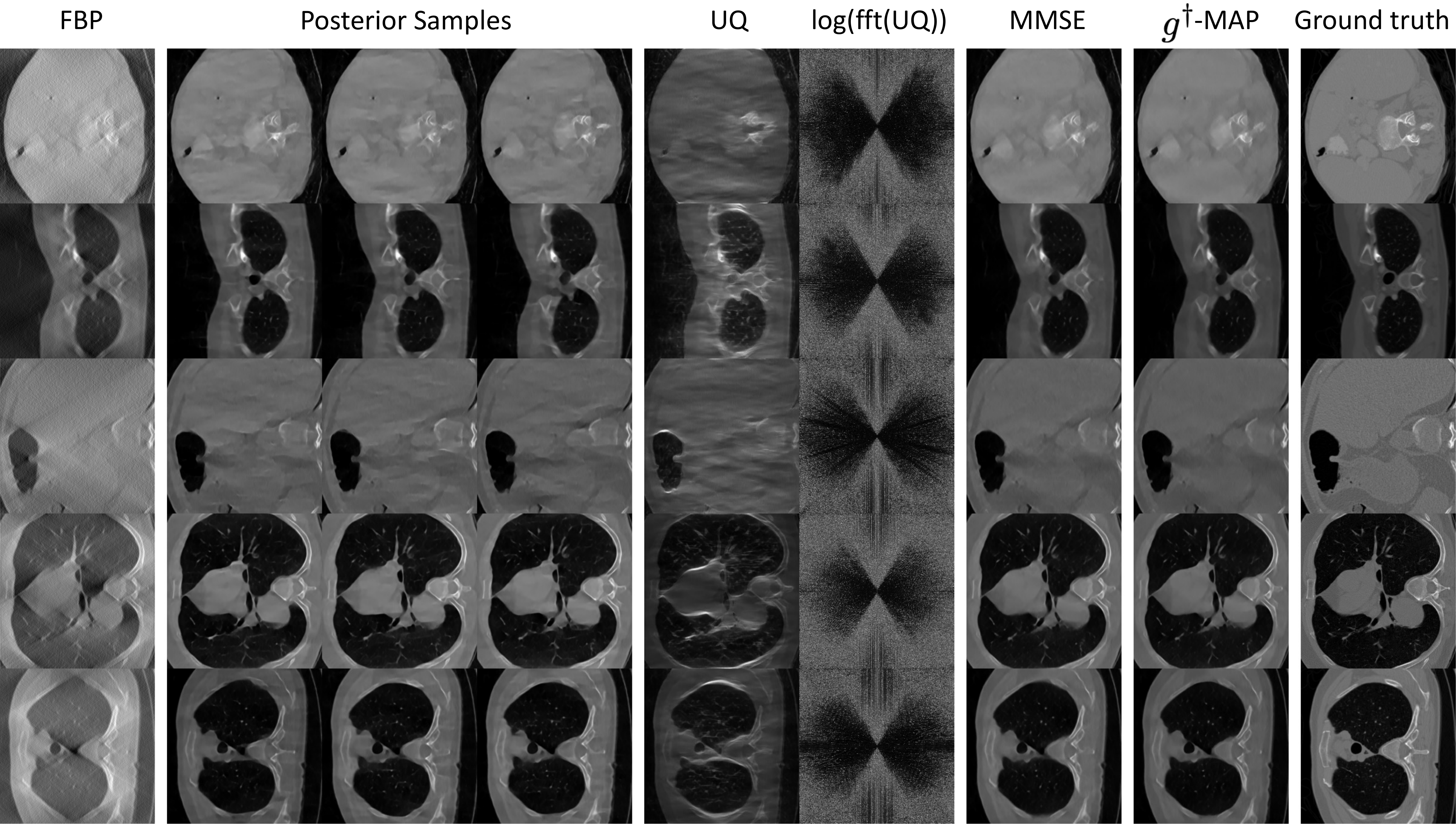}
\caption{Vertical missing cone ($60^\circ$ to $120^\circ$)}
\end{subfigure}
\begin{subfigure}{\textwidth}
\centering
\includegraphics[width=0.9\textwidth]{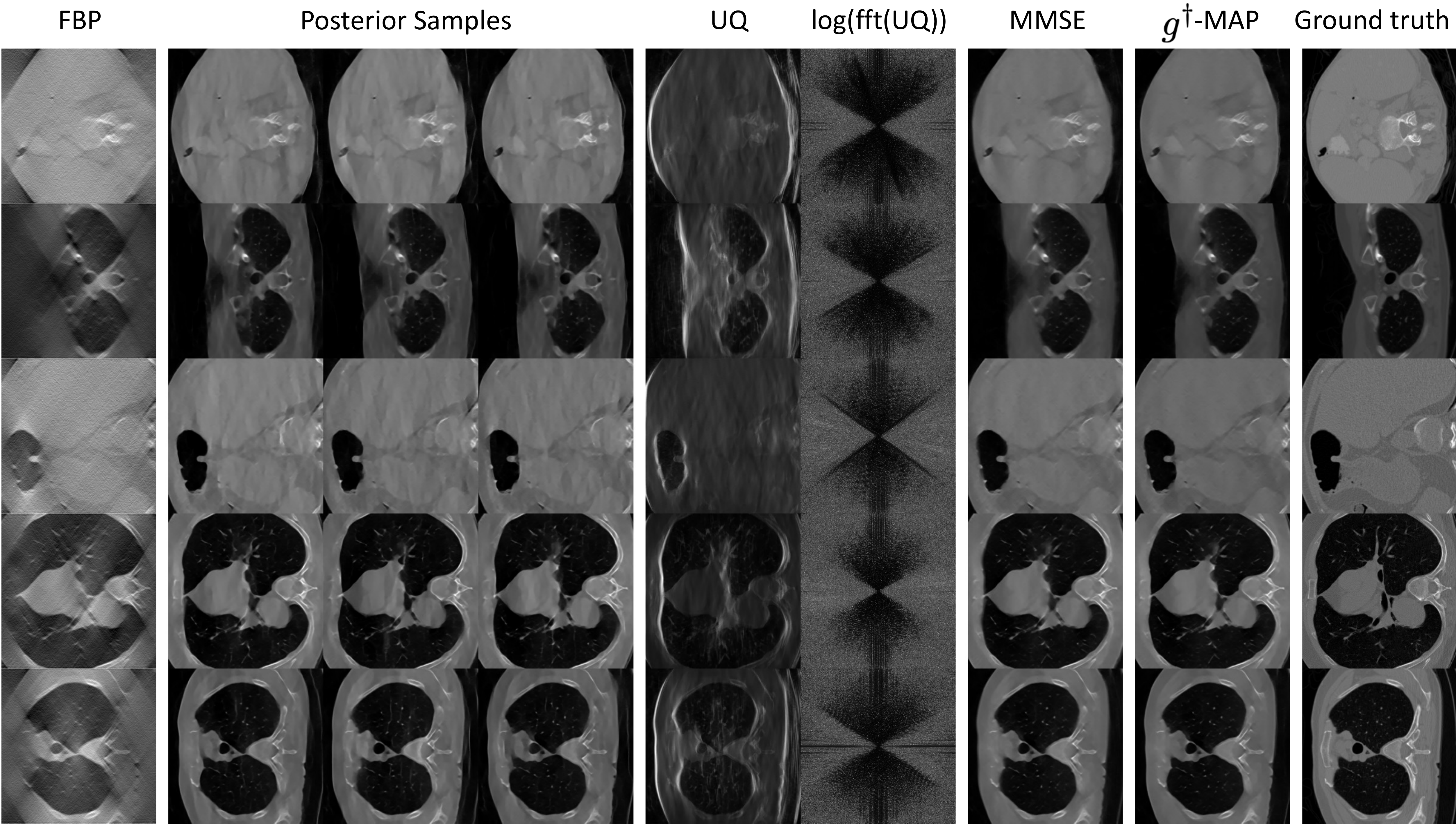}
\caption{Horizontal missing cone($-30^\circ$ to $30^\circ$)}
\end{subfigure}
\caption{Limited-view CT in resolution $256 \times 256$}
\label{fig:Limited-CT_256 samples}
\end{figure*}

\begin{figure*}
\centering
       \includegraphics[width=0.7\textwidth]{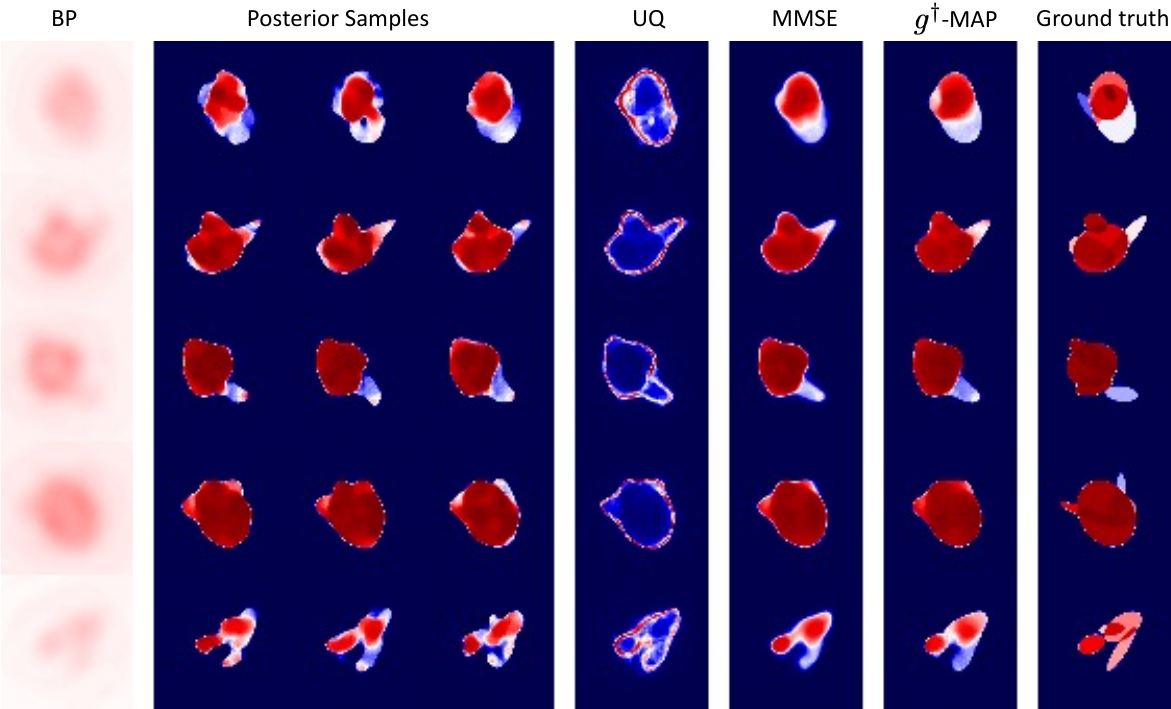}
\caption{Inverse scattering ($\epsilon_r=1.5$,  full-view)}
\label{fig:scattering_samples_er1.5}
\end{figure*}

\begin{figure*}
\centering
\includegraphics[width=0.7\textwidth]{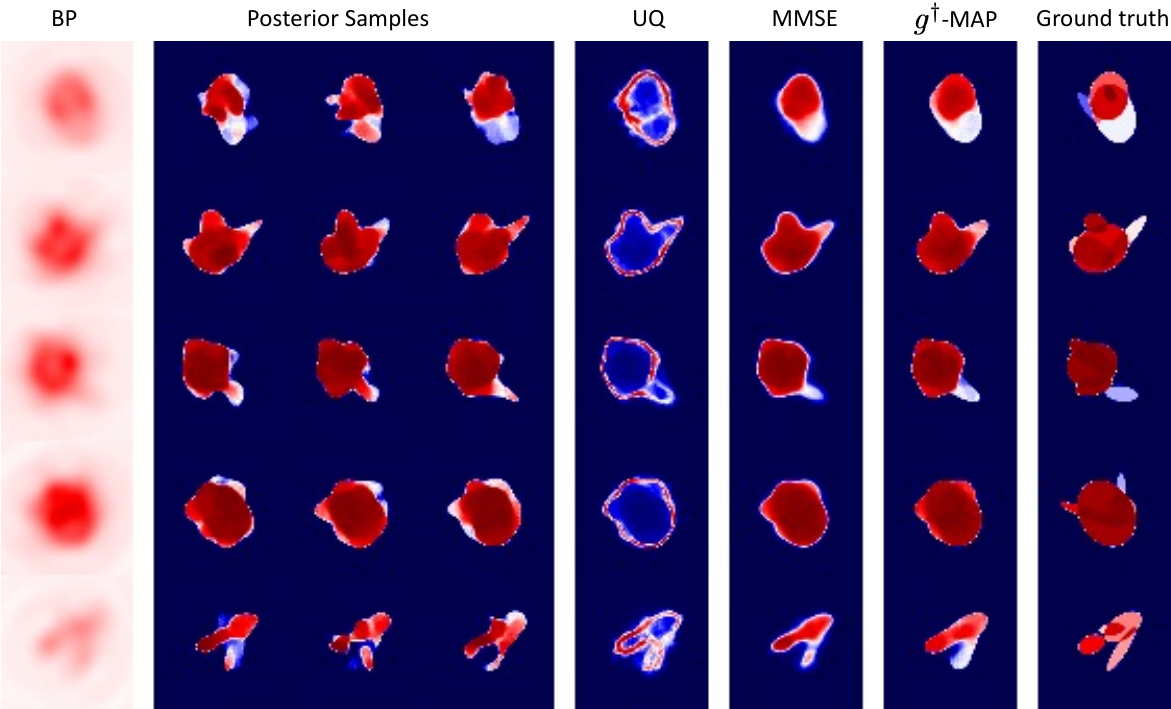}
\caption{Inverse scattering ($\epsilon_r=2$,  full-view)}
\label{fig:scattering_samples_er2}
\end{figure*}

\begin{figure*}
\centering
\includegraphics[width=0.7\textwidth]{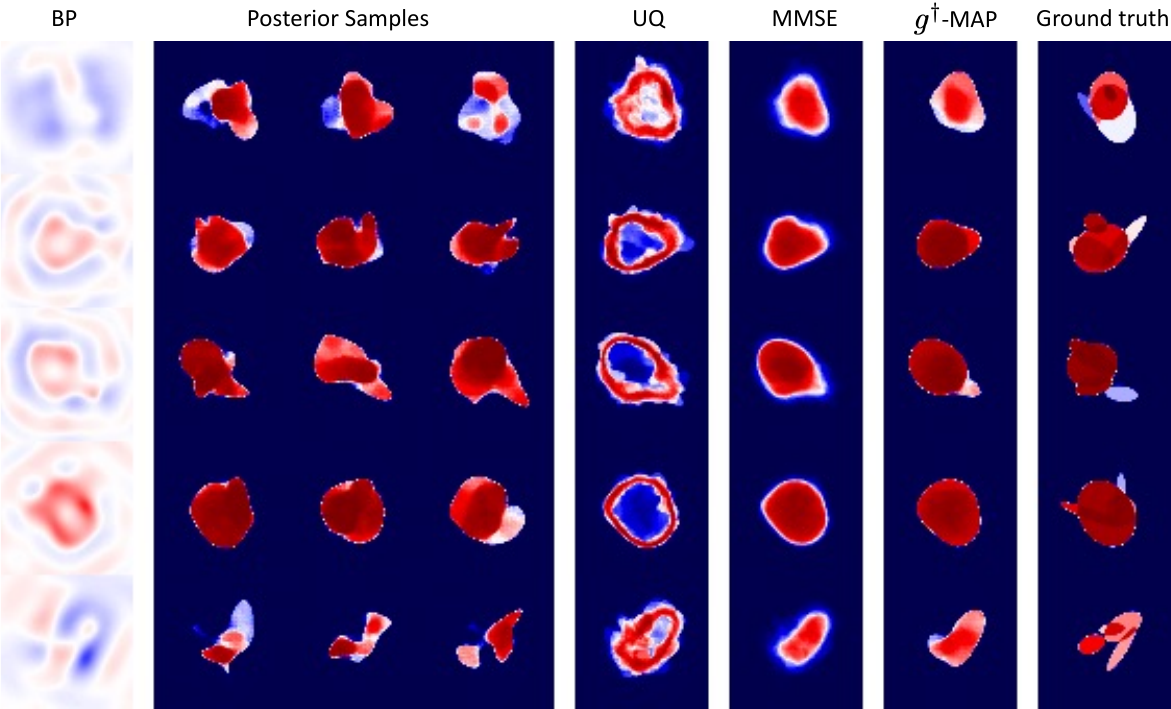}
\caption{Inverse scattering ($\epsilon_r=6$,  full-view)}
\label{fig:scattering_samples_er6}
\end{figure*}

\begin{figure*}
\centering
\includegraphics[width=0.7\textwidth]{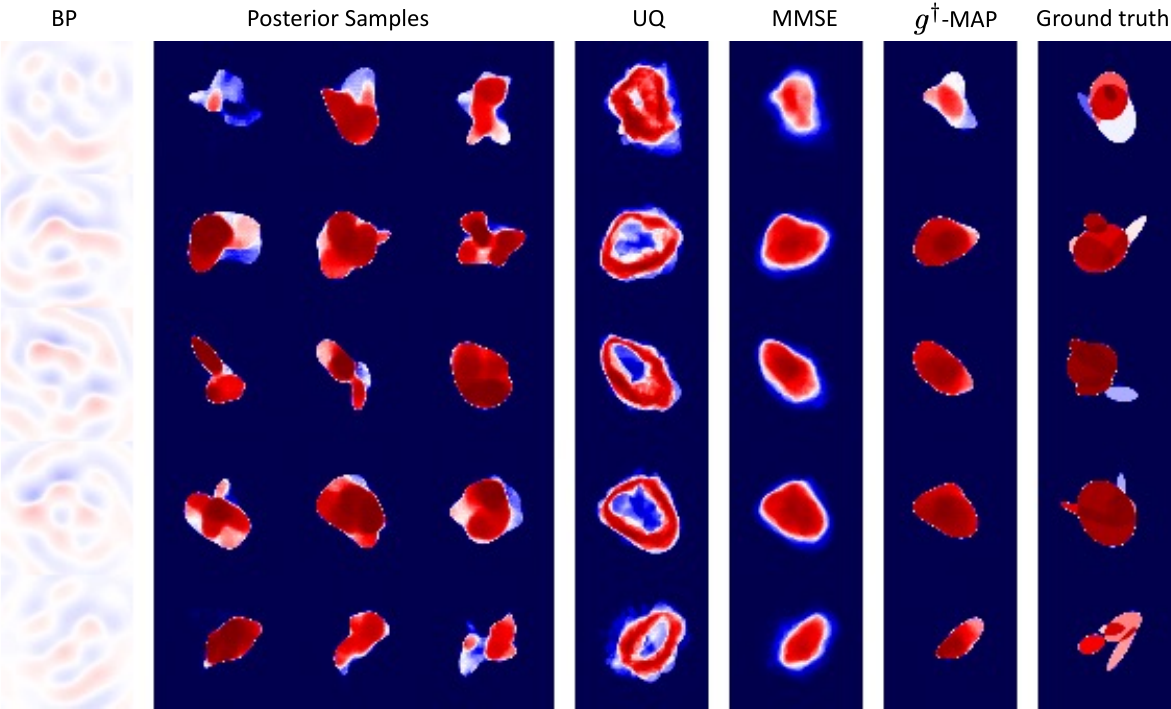}
\caption{Inverse scattering ($\epsilon_r=6$,  top-view)}
\label{fig:scattering_samples_slice_er6}
\end{figure*}

\begin{figure*}
\centering
\includegraphics[width=0.7\textwidth]{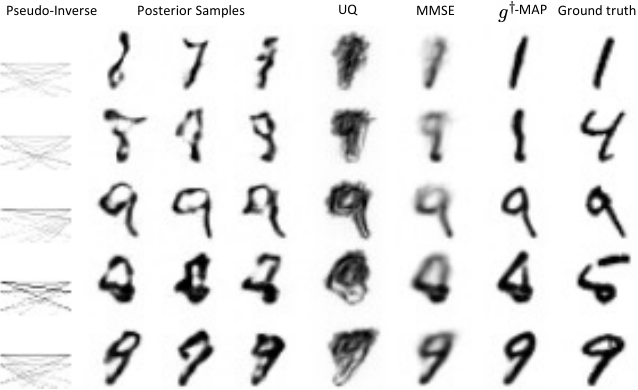}
\caption{Travel-time tomography ($NS = 10$)}
\label{fig:traveltime_4}
\end{figure*}

\begin{figure*}
\centering
\includegraphics[width=0.7\textwidth]{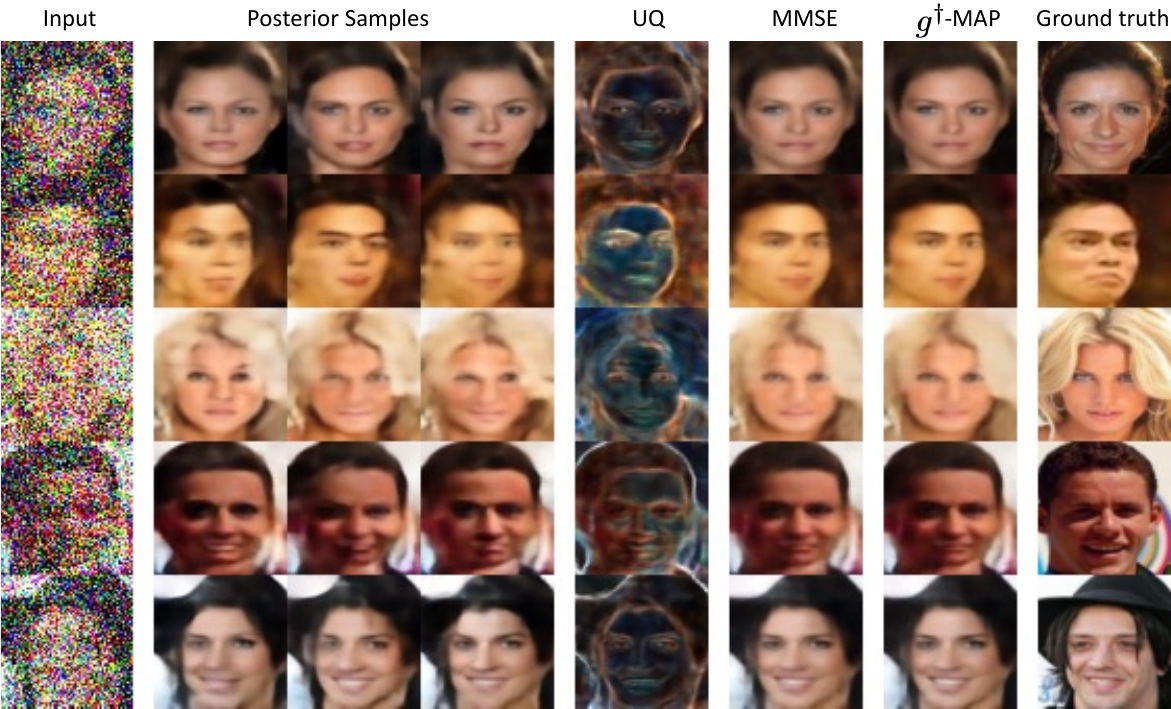}
\caption{Denoising}
\label{fig:denoising_m1_samples}
\end{figure*}

\begin{figure*}
\centering
\includegraphics[width=0.7\textwidth]{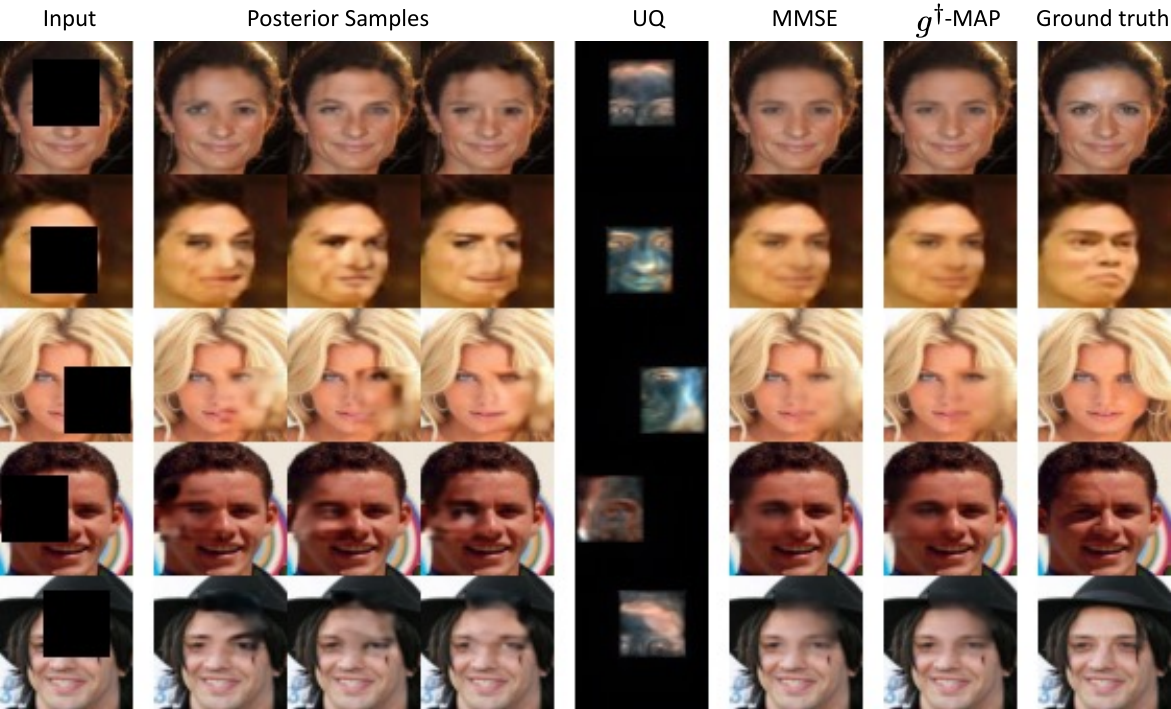}
\caption{Mask ($s=32$)}
\label{fig:mask32_samples}
\end{figure*}

\begin{figure*}
\centering
\includegraphics[width=0.7\textwidth]{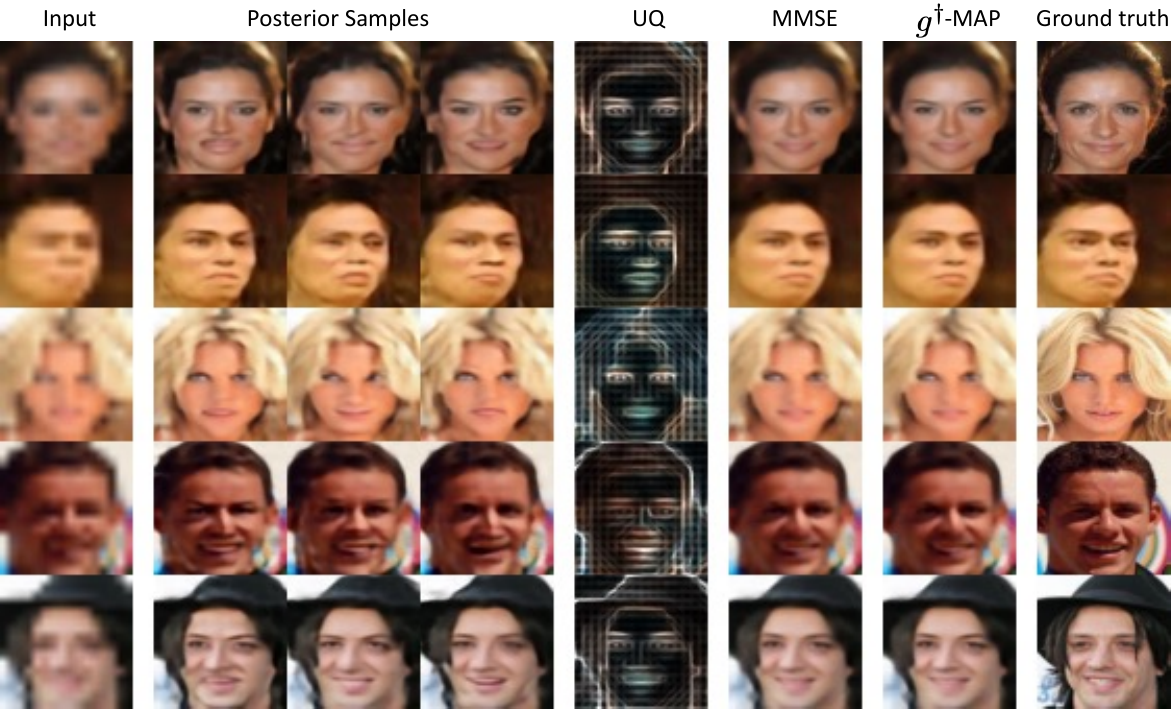}
\caption{Super resolution ($f=4$)}
\label{fig:sr4_samples}
\end{figure*}

\begin{figure*}
\centering
\includegraphics[width=0.7\textwidth]{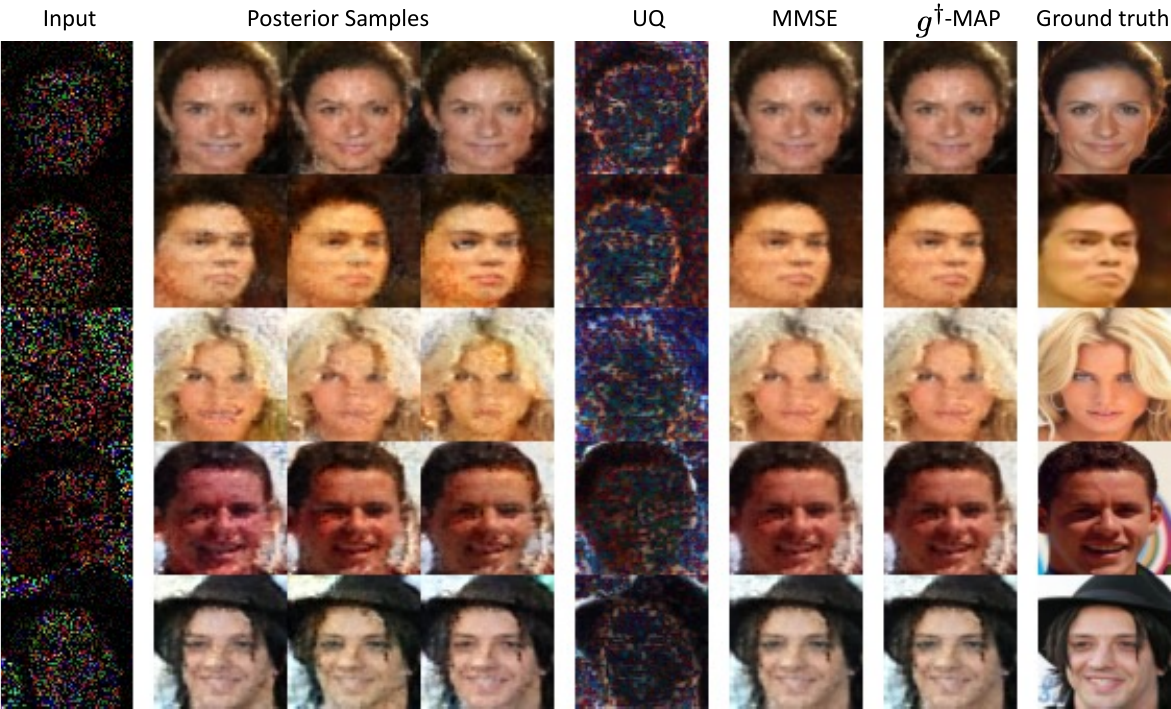}
\caption{Random mask ($p=0.2$)}
\label{fig:random_mask_samples}
\end{figure*}

Figure~\ref{fig: Compressed sensing comparison} compares the performance of conditional normalizing flows in image restoration tasks. Figures~\ref{fig:Limited-CT_256 samples} to ~\ref{fig:random_mask_samples}    demonstrate further results on different ill-posed inverse problems. We can observe that C-Trumpets have a significant edge over the baselines. 

\vfill
\end{document}